\documentclass{article}

\usepackage[preprint]{corl_2026} 

\usepackage{enumitem}
\usepackage{graphicx}
\usepackage{xspace}
\usepackage{amsmath}
\usepackage{amssymb}
\usepackage{url}
\usepackage{setspace} 
\usepackage{algorithm}
\usepackage{algpseudocode}

\usepackage{multirow}
\usepackage{tcolorbox}
\usepackage{tabularx}
\usepackage{booktabs}
\usepackage{makecell}
\usepackage{array}
\usepackage{utfsym}
\usepackage{caption}
\usepackage[dvipsnames]{xcolor}
\newcommand{\coloredcheckmark}[1]{\textcolor{#1}{\usym{2714}}}
\newcommand{\coloredcross}[1]{\textcolor{#1}{\usym{2718}}}
\definecolor{ForestGreen}{rgb}{0.1333,0.545,0.1333}
\definecolor{Firebrick}{rgb}{0.698,0.1333,0.1333}
\definecolor{DeepGreen}{RGB}{127,147,92}
\definecolor{DeepBlue}{RGB}{37,83,139}

\newcolumntype{Y}{>{\centering\arraybackslash}X}
\usepackage[table]{xcolor}
\newcommand{\minisection}[1]{\noindent{\textbf{#1.}}}
\newcommand{\tablestyle}[2]{\setlength{\tabcolsep}{#1}\renewcommand{\arraystretch}{#2}\centering\footnotesize}
\newlength\savewidth
\usepackage{subcaption}
\newcommand{\Ours}{T-Rex\xspace}

\title{\Ours: Tactile-Reactive Dexterous Manipulation}

\author{
  \parbox{\linewidth}{\centering
    Dantong Niu\textsuperscript{1,2*}, Zhuoyang Liu\textsuperscript{1*}, Zekai Wang\textsuperscript{1*}, Boning Shao$^{1}$, \\[0.01in] Zhao-Heng Yin$^{1}$,  Anirudh Pai$^{1}$, Yuvan Sharma$^{1}$, Stefano Saravalle$^{5}$, Ruijie Zheng$^{2}$,  \\[0.01in]  Jing Wang$^{2}$,   Ryan Punamiya$^{2}$, Mengda Xu$^{2}$,  Yuqi Xie$^{2}$, Yunfan Jiang$^{2,3}$, Letian Fu$^{1}$, \\[0.01in] 
    Konstantinos Kallidromitis$^{4}$, Matteo Gioia$^{5,6}$, Junyi Zhang$^{1}$, Jiaxin Ge$^{1}$,  Haiwen Feng$^{1}$, \\[0.01in]
    Fabio Galasso$^{5,6}$, Wei Zhan$^{1}$, David M. Chan$^{1}$, Yutong Bai$^{1}$, Roei Herzig$^{1}$, Jiahui Lei$^{1}$, \\[0.01in]
    Li Fei-Fei$^{3}$, Ken Goldberg$^{1}$, Jitendra Malik$^{1}$, Pieter Abbeel$^{1}$, Yuke Zhu$^{2}$, Danfei Xu$^{2}$, Jim(Linxi) Fan$^{2}$, Trevor Darrell$^{1}$ \\[0.06in]
    \small
    $^1$UC Berkeley \qquad $^2$NVIDIA \qquad $^3$Stanford \qquad  $^4$Panasonic \qquad $^5$La Sapienza University \qquad $^6$ItalAI \\[0.05in]
    $^*$Equal Contribution \\[0.05in]
    \texttt{\color{DeepBlue}https://tactile-rex.github.io/} 
  }
}

\begin{document}
\maketitle

\vspace*{-1.3cm}
\begin{center}
    \centering
    \includegraphics[width=0.95\linewidth]{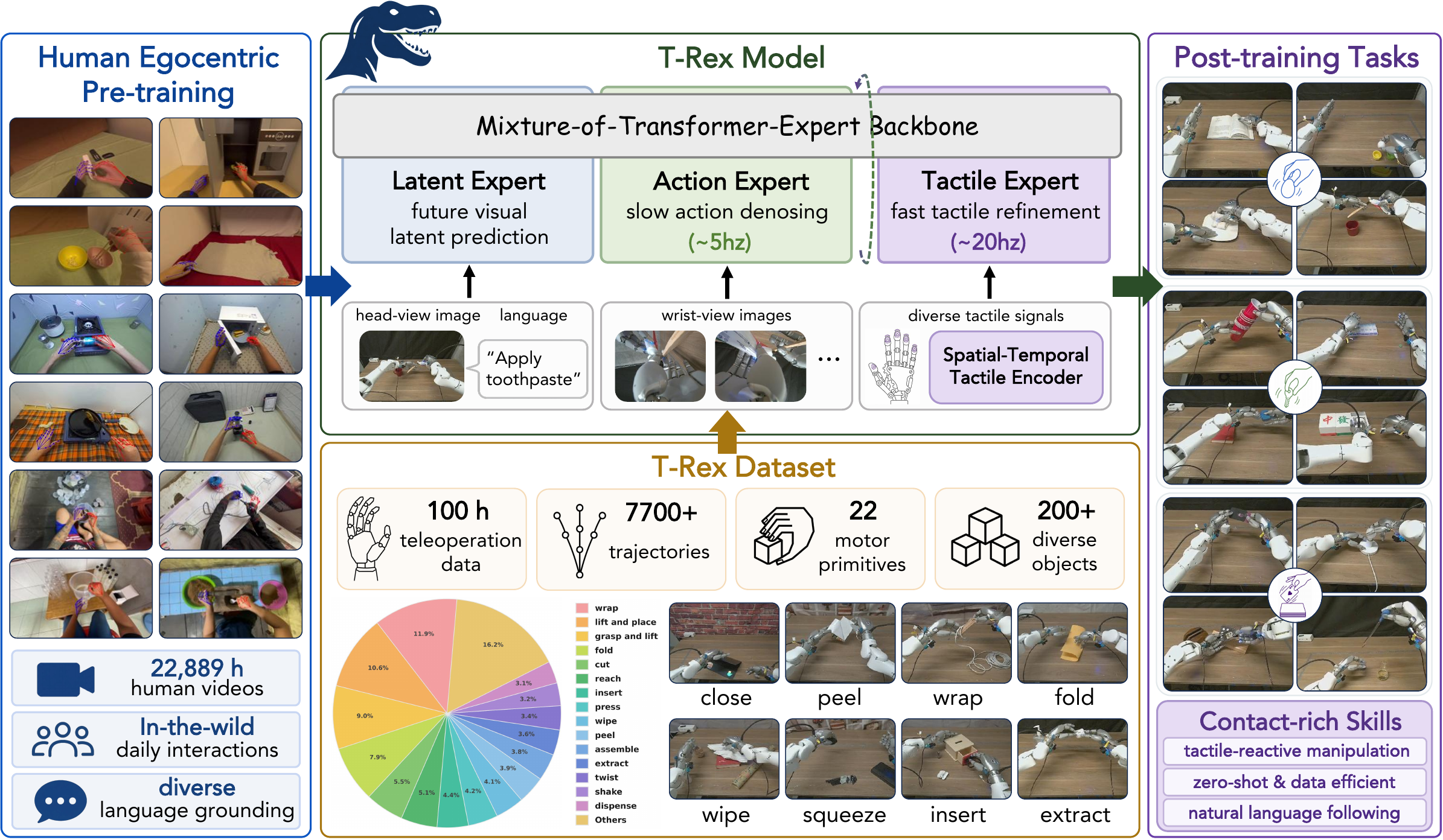}
    \vspace{0.1Cm}
    \captionof{figure}{\textbf{Overview}. T-Rex is a tactile-reactive dexterous manipulation framework that combines large-scale human egocentric pre-training with tactile-grounded robot mid-training. Built on a Mixture-of-Transformer-Experts (MoT) architecture, the T-Rex Model integrates low-frequency visuomotor planning with high-frequency tactile refinement and employs a spatial-temporal tactile encoder. The T-Rex Dataset introduces 100 hours of real robot data containing diverse motor primitives and object interactions with synchronized tactile signals as well as 12 manipulation tasks.}\label{fig:teaser}
\end{center}

\begin{abstract}
The ability to react dynamically to tactile signals has long been considered crucial to agile human-level dexterity. Yet contemporary learning-based VLAs for robotic manipulation generally either overlook the tactile modality or are limited to encoders with static cues, in part due to the scarcity of diverse training data and standardized evaluation, architectural constraints in current Vision-Language-Action (VLA) models, and limitations of static tactile encoders. In this paper, we push the frontier of tactile-reactive manipulation addressing all of these limitations. We {propose} a large-scale, 100-hour tactile-rich dataset collected via a novel, data-efficient recipe that prioritizes elementary motor primitives. To effectively exploit naturally high-frequency touch signals without sacrificing the existing capabilities of existing VLAs, we introduce a variable-rate Mix-of-Transformer (MoT) architecture equipped with a novel temporal tactile VQ-VAE encoder. We demonstrate the effectiveness of tactile-reactive policies on 12 manipulation tasks requiring delicate force control, deformable object manipulation, achieving over 30\% higher average success rate than the strongest baseline.

\end{abstract}

\keywords{Tactile, Haptic, Dexterous Manipulation, Contact-rich Dataset} 
\newcommand{\dantong}[1]{{\color{violet}\textbf{[DN:} #1]}}

\vspace{-1em}
\section{Introduction}
\label{sec:intro}
\vspace{-0.2cm}

Human dexterity relies on more than vision; it depends fundamentally on the ability to feel and rapidly react to fine-grained tactile signals. 
While everyday tasks like sliding a thin card into a slot or opening a lock with a key are effortless for humans, they remain challenging for current robot learning policies. 
Mastering them requires \textit{tactile-reactive} behaviors: immediate, closed-loop motor responses to tactile signals, far faster than conventional vision-based control loops allow.

However, adapting existing Vision-Language-Action (VLA) models for tactile-reactive manipulation presents two major challenges. First, tactile datasets for policy pre-training remain scarce. As a result, existing large-scale pre-training paradigms~\cite{zheng2026egoscale,yang2025egovla,kareer2024egomimic,tao2025dexwild} are predominantly visual, missing critical physical signals such as force variations, micro-slips, and local deformations. 
Learning such policies from scratch would require massive amounts of synchronized visuo-tactile data—a scale at which collecting fine-grained dexterous teleoperation becomes prohibitively expensive.
To overcome this, we demonstrate that tactile capabilities can be efficiently acquired during a dedicated mid-training phase, bypassing the need for tactile data during pre-training. To this end, we introduce the \textbf{\Ours Dataset}, a 100-hour, tactile-synchronized teleoperation dataset.
Rather than recording narrow, task-specific demonstrations, we design it around diverse verb-noun combinations, covering contact-rich behaviors through compositional motor primitives and object interactions. 
As illustrated in Fig.~\ref{fig:teaser}, we formulate a training paradigm that distills general visuomotor priors from large-scale human video pre-training, aligns interaction dynamics via tactile-rich mid-training, and rapidly adapts to downstream tasks with minimal target-domain demonstrations.


Beyond the data bottleneck, significant architectural challenges remain due to the fundamental frequency mismatch: tactile-reactive control requires high-frequency responses, whereas standard VLM backbones operate at lower frequencies. 
While recent dual-system architectures attempt to address this by completely separating fast motor responses from cognitive reasoning~\cite{black2024pi0,bjorck2025gr00t,chen2025fastinslowdualsystemfoundationmodel}, and variable-rate diffusion policies remain confined to task-specific imitation learning for parallel grippers~\cite{xue2025reactive}, we propose a unified foundation model for dexterous manipulation. 
The \textbf{\Ours Model} is a multi-modal framework built upon a variable-rate Mixture-of-Transformers (MoT) that disentangles control into a low-rate action expert for baseline dexterous manipulation and a high-rate tactile expert for rapid residual refinements. 
This high-rate expert relies on a spatio-temporal VQ-VAE to capture fine-grained tactile dynamics, compressing raw tactile feedback into compact representations of force and contact. 
To align this tactile modality with the broader visual context during mid-training, we employ an auxiliary objective for future visual latent prediction.
Together, these components extend standard VLA capabilities with high-frequency, closed-loop tactile behaviors.


To evaluate tactile-reactive control, we introduce a real-world benchmark of 12 contact-rich dexterous manipulation tasks spanning insertion, deformation, force-sensitive interaction, and bimanual coordination. Across these tasks, \Ours improves average success rate by 30\% over existing dexterous-hand foundation models, with stronger robustness and generalization in contact-rich manipulation.
In summary, our contributions are threefold: (i) the \textbf{\Ours Dataset}, an open-source 100-hour tactile-synchronized teleoperation dataset organized around motor primitives and object interactions; (ii) the \textbf{\Ours Model}, a variable-rate Mixture-of-Transformer (MoT) with a spatio-temporal tactile VQ-VAE, trained via our mid-training recipe for high-frequency closed-loop control; and (iii) a real-world evaluation benchmark for tactile-reactive dexterous manipulation on which \Ours establishes a strong baseline for future work.

\vspace{-1em}
\vspace{0.2cm}
\section{Related Work}
\label{sec:rel}

\minisection{Tactile Sensing for Manipulation} Leveraging tactile sensing for robot manipulation has received increasing attention, with prior works exploring tactile representations and fusion architectures \citep{fu2024tvl, sferrazza2023power, zhu2025touch, guzey2023dexterity, gao2016tactileunderstanding, liu2026mlamultisensorylanguageactionmodel}. Early methods injected tactile observations into imitation learning policies with simple models such as shallow MLPs \citep{lin2024learning}, while later work introduced more structured tactile modeling, including rigid-body-pose-aware encodings \citep{Wu2024CanonicalRA, Huang2025SpatiallyAT}, joint prediction of future visual or tactile observations and actions \citep{heng2025vitacformer, vtam2026}, and tactile-aware low-level action refinement \citep{xue2025reactive, yu2024mimictouch}. Recent VLA models incorporate touch by treating tactile signals as an additional modality \citep{Huang2025TactileVLA, Zhang2025VTLAVM}, aligning tactile representations with visual and language latents \citep{cheng2025omnivtla, jones24fuse}, or designing architectures that better exploit force and contact signals \citep{bi2025vlatouch, yu2025forcevla}. Despite this progress, there is still no canonical tactile-aware recipe that benefits from large-scale pretraining and midtraining for dexterous manipulation, and existing work largely focuses on single-arm or parallel-gripper embodiments. Our work builds on these efforts with a tactile-reactive MoT architecture featuring an asynchronous high-frequency tactile expert for challenging contact-rich dexterous bimanual manipulation.

\minisection{Unified Multimodal Models} Recent foundation models increasingly adopt unified architectures that jointly model multiple modalities, including text, images, video, and actions \citep{deng2025emergingpropertiesunifiedmultimodal, chameleonteam2024, chen2025janus}. In robot learning, early VLA models directly finetune VLM backbones to generate action tokens \citep{zitkovich2023rt, kim2024openvla}, while recent approaches introduce dedicated action experts for continuous action generation \citep{black2024pi0, bjorck2025gr00t}. Several works further extend VLAs with future prediction \citep{hu2024vpp, lv2025f1, cheang2024gr2}, world-modeling experts \citep{cai2026internvlaa1unifyingunderstandinggeneration, bi2025motusunifiedlatentaction, hu2026bagelvlaenhancinglonghorizonmanipulation}, or additional reasoning modules \citep{huang2025motvlavisionlanguageactionmodelunified, gu2025manualvlaunifiedvlamodel, zhao2025cotvla, liu2026last0latentspatiotemporalchainofthought}. Our work expands on the previous latent prediction plus action generation experts design by fusing tactile information as a novel modality through asynchronous MoT, unlocking potentials for tactile-reactive manipulation.

\minisection{Egocentric Human Videos in Dexterous Manipulation} Large-scale human egocentric manipulation datasets \citep{hoque2026egodex, Grauman2022Ego4D, Liu2022HOI4D} have motivated robot learning from human demonstrations. Prior work leverages human video for self-supervised representation learning \citep{Xiao2022MVP, nair2022r3m, niu2025arm4r}, affordance extraction \citep{kannan2023deft, li2025affgrasp}, VLA pretraining \citep{yang2025egovla}, human-robot co-training \citep{kareer2024egomimic, tao2025dexwild}, World Action Models \citep{gao2026dreamdojo, ye2026dreamzero, li2025uvam}, hierarchical planning and control \citep{xu2025flow, wang2023mimicplay}, and human-to-robot transfer \citep{zheng2025flare, kareer2025emergencehumanrobottransfer}. EgoScale \citep{zheng2026egoscale} further studies scaling laws for manipulation policies pretrained on human egocentric data. Following these works, we leverage large-scale human egocentric pretraining to provide T-Rex with broad visuomotor priors, and further align the model through large-scale teleoperated mid-training data with synchronized tactile feedback, enabling contact-rich and tactile-reactive behaviors.


\section{The \Ours Dataset }


\begin{figure}[h]
    \centering
    \includegraphics[width=\linewidth]{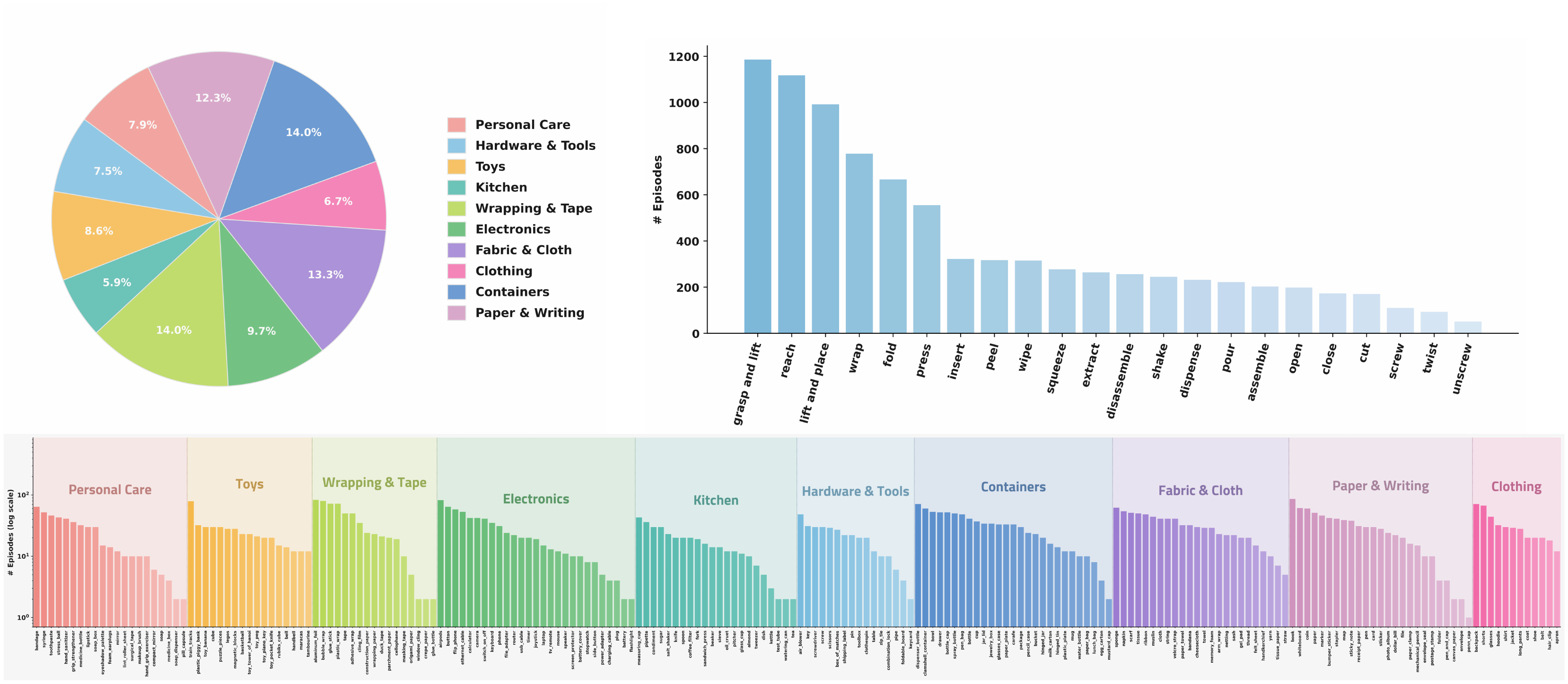}
    \caption{\textbf{Statistics of the T-Rex Dataset.} Top-left: distribution of object categories. Top-right: distribution of motor primitives. Bottom: long-tail distribution of individual objects. \Ours dataset contains 100 hours of tactile-synchronized bimanual manipulation mid-training data spanning 200+ daily objects and 22 motor primitives, designed for broad coverage of contact-rich interactions.}
    \label{fig:play_data_diversity}
    \vspace{-0.4cm}
\end{figure}

Existing robot manipulation datasets primarily focus on parallel-gripper systems or grasp-centric dexterous hands~\citep{oxe2023, khazatsky2024droid, walke2023bridgedata, dasari2019robonet, lim2026HRDexDB, liu2024realdex, wang2022dexgraspnet, zhang2024graspxl, wang2025dexh2r, fan2023arctic}, offering limited coverage of tactile-rich dexterous interactions. To support tactile-reactive policy learning, we collect a 100-hour bimanual dexterous manipulation dataset spanning over 200 everyday objects and 22 motor primitives, covering a diverse set of contact-rich behaviors. Each episode contains synchronized RGB observations, tactile signals, robot states, actions, and language instructions. Fig.~\ref{fig:play_data_diversity} and App.~\ref{apdx:trex_dataset} shows the distribution and details.

\section{Tactile-Reactive
Dexterous Manipulation}

\label{sec:method}



The \Ours policy $\pi_\theta$ receives RGB observations $\mathbf{o}_t$, language instructions $\ell$, tactile force history $\mathbf{f}_{t-H_f:t}$, and tactile deformation maps $\mathbf{d}_t$. Denoting the multimodal context as $\mathbf{c}_t=\{\mathbf{o}_t,\ell,\mathbf{f}_{t-H_f:t},\mathbf{d}_t\}$, the policy predicts a future action chunk $\mathbf{A}_{t:t+H}$ with a horizon $H$. Following standard flow-based robot policies, action generation is formulated as conditional flow matching. 
The model learns a vector field \(v_\theta(x_\tau,\tau \mid c_t)\) and is trained using Loss Eq. (1), where \(x_0 = A_{t:t+H}\) denotes the clean action chunk and \(x_1 = \epsilon \sim \mathcal{N}(0,I)\) denotes Gaussian noise.

\vspace{-1em}
\begin{equation}
\mathcal{L}_{\mathrm{FM}}(\theta)
=
\mathbb{E}
\left[
\left\|
v_\theta(x_\tau,\tau \mid c_t) - (x_1 - x_0)
\right\|^2
\right]
\label{eq:flow_matching}
\end{equation}


\subsection{Model Architecture}
\label{sec:model_architecture}

\begin{figure}[!t]
    \centering
    \includegraphics[width=\linewidth]{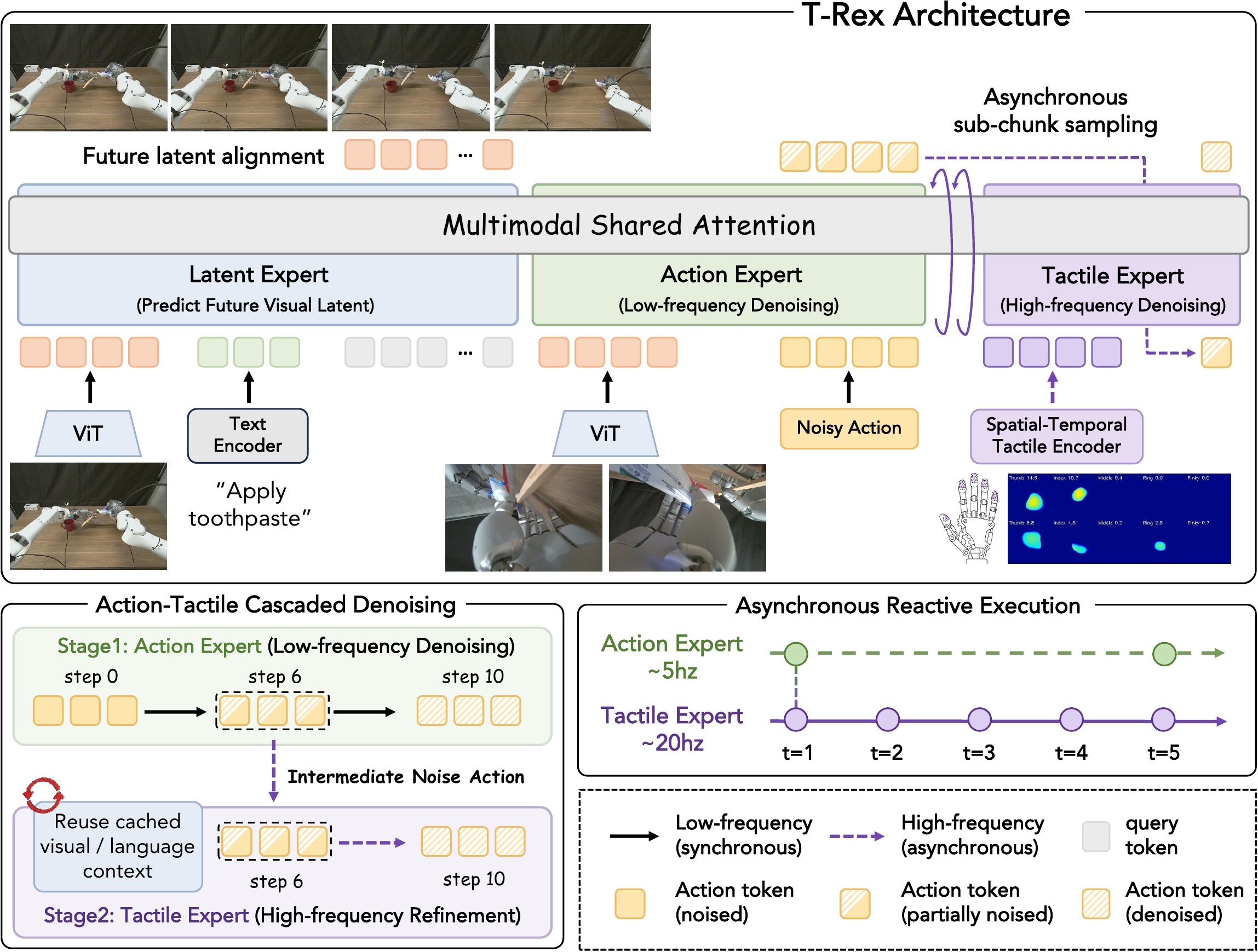}
    \vspace{0.1cm}
    \caption{\textbf{T-Rex Model Architecture.} T-Rex uses a Mixture-of-Transformer-Experts (MoT) backbone with three experts: a latent expert for future visual prediction, an action expert for low-frequency action denoising, and a tactile expert for high-frequency tactile refinement. During inference, the tactile expert reuses cached visual-language context to asynchronously refine intermediate actions using spatial-temporal tactile features, enabling fast tactile-reactive closed-loop control.}
    \label{fig:model_arch}
    \vspace{-0.1cm}
\end{figure}

As shown in Fig.~\ref{fig:model_arch}, \Ours consists of a Mixture-of-Transformer-Experts (MoT) backbone and a spatial-temporal tactile encoder and that together enable tactile-reactive dexterous manipulation.

\minisection{Mixture-of-Transformer-Experts Backbone}
\Ours employs a Mixture-of-Transformer-Experts (MoT) backbone with three specialized experts. The \textbf{latent expert} processes visual and language observations to predict future visual representations, providing temporally grounded context. The \textbf{action expert} generates a low-frequency action plan by denoising actions from pure noise to an intermediate timestep $\tau_{\mathrm{split}}$. The \textbf{tactile expert} then reuses the cached visual-language context and continues denoising from $\tau_{\mathrm{split}}$ to $\tau=0$, refining the action using high-frequency tactile observations to produce the final executable action chunk $\mathbf{A}_{t:t+H}$. Implementation details are provided in App.~\ref{apdx:model-training}.

\minisection{Spatial-Temporal Tactile Encoding}
\label{sec:tactile_encoding}
We encode tactile observations using both temporal force dynamics and spatial deformation signals. A per-finger VQ-VAE compresses the recent force history $\mathbf{f}_{t-15:t}$ into compact temporal tokens, while the current force vector $\mathbf{f}_t$ is projected directly to preserve instantaneous contact information. In parallel, a convolutional encoder extracts features from the current deformation map $\mathbf{d}_t$. The resulting features are concatenated to form the tactile token sequence $\mathbf{z}^{\tau}_t$ in Eq.~\eqref{eq:tactile_encoding}. The implementation details of VQ-VAE and tactile encoder refers to App.~\ref{apdx:tacile_encoder}.

\begin{equation}
\mathbf{z}^{\tau}_t
=
\bigl[
\mathrm{Emb}_{\mathrm{vq}}\!\bigl(E_f(\mathbf{f}_{t-15:t})\bigr);
\mathrm{Proj}_f(\mathbf{f}_t);
\mathrm{Proj}_d\!\bigl(E_d(\mathbf{d}_t)\bigr)
\bigr].
\label{eq:tactile_encoding}
\end{equation}


\subsection{Asynchronous Tactile-Reactive Cascaded Flow Matching}
\label{sec:cascaded_flow}

\minisection{Asynchronous Tactile-Reactive Cascaded Flow Matching} To combine low-frequency visuomotor planning with high-frequency tactile refinement, we split the flow-matching trajectory at a fixed timestep $\tau_{\mathrm{split}}$. The action expert denoises the upper segment $\tau\in[\tau_{\mathrm{split}},1]$ to produce an intermediate action state $\hat{\mathbf{x}}_{\tau_{\mathrm{split}}}$, while the tactile expert reuses the cached visual-language context and completes the remaining denoising process $\tau\in[0,\tau_{\mathrm{split}}]$ using real-time tactile observations.


\minisection{Shared Flow Target}
Given a demonstrated action chunk $\mathbf{A}^{\mathrm{demo}}$ and Gaussian noise $\boldsymbol{\epsilon} \sim \mathcal{N}(\mathbf{0}, \mathbf{I})$, the linear interpolant $\mathbf{x}_\tau$ and constant velocity target $v^\star$ are defined in Eq.\eqref{eq:asyn}. Both experts regress this identical target $v^\star$ over disjoint sub-intervals of $\tau \in (0, 1]$, conditioned on global multimodal contexts (upper segment) and localized tactile observations (lower segment), respectively.

\begin{equation}
\mathbf{x}_\tau = (1-\tau)\,\mathbf{A}^{\mathrm{demo}} + \tau\,\boldsymbol{\epsilon},
\qquad
v^\star = \boldsymbol{\epsilon} - \mathbf{A}^{\mathrm{demo}}.
\label{eq:asyn}
\end{equation}

\minisection{Cascaded Denoising Inference}
During inference, we use $N=10$ of Euler steps and split at $\tau_{\mathrm{split}} = 0.4$. The \textbf{slow stream} from action expert runs once per action chunk, integrating the upper trajectory from $\mathbf{x}_1 = \boldsymbol{\epsilon}$ over $K_{\mathrm{slow}} = 6$ steps ($\Delta\tau = -0.1$):
\begin{equation}
\hat{\mathbf{x}}_{\tau_{\mathrm{split}}} = \mathrm{Euler}\bigl(f_\theta^{\mathrm{act}};\, \mathbf{x}_1,\, \tau{:}\,1{\to}0.4,\, K_{\mathrm{slow}}=6\bigr).
\end{equation}
After this partial denoising, the boundary states are cached as a stationary visual context $\mathrm{KV}_{\tau_{\mathrm{split}}}$. For an action chunk length of $T_a=16$, the \textbf{fast stream} from the tactile expert is triggered frequently at offsets $\{0, 4, 8, 12\}$ within the chunk, which bypasses the heavy visual network, feeding real-time tactile tokens $\mathbf{z}^{\tau}_t$ and $\mathrm{KV}_{\tau_{\mathrm{split}}}$ into the tactile expert to resolve the remaining $K_{\mathrm{fast}} = 4$ steps:
\begin{equation}
\mathbf{A}_{t:t+T_a} = \mathrm{Euler}\bigl(f_\theta^{\mathrm{tac}};\, \hat{\mathbf{x}}_{\tau_{\mathrm{split}}},\, \tau{:}\,0.4{\to}0,\, K_{\mathrm{fast}}=4\bigr).
\end{equation}
This mechanism enables the model to respond dynamically to real-time tactile feedback, facilitating the execution of contact-rich and highly dexterous tasks.

\minisection{Training Protocol}
During training, we sample $\tau_{\mathrm{act}} \sim \mathrm{Beta}(1.5, 1.0)$ on $(0, 1]$ for the action expert, and $\tau_{\mathrm{tac}} = \tau_{\mathrm{split}} \cdot \tilde{\tau}$ where $\tilde{\tau} \sim \mathrm{Beta}(1.5, 1.0)$ on $(0, \tau_{\mathrm{split}}]$ for the tactile expert. Both networks jointly minimize their respective mean-squared error against the shared target $v^\star$:
\begin{equation}
\mathcal{L}_{\mathrm{act}} = \bigl\lVert f_\theta^{\mathrm{act}}(\mathbf{x}_{\tau_{\mathrm{act}}}, \tau_{\mathrm{act}}) - v^\star \bigr\rVert^2,
\quad
\mathcal{L}_{\mathrm{tac}} = \bigl\lVert f_\theta^{\mathrm{tac}}(\mathbf{x}_{\tau_{\mathrm{tac}}}, \tau_{\mathrm{tac}};\, \mathrm{KV}_{\tau_{\mathrm{split}}}) - v^\star \bigr\rVert^2,
\end{equation}
where $\mathrm{KV}_{\tau_{\mathrm{split}}}$ is extracted from a detached slow-stream pass. Notably, training the action expert across the full $(0, 1]$ domain ensures it retains standalone competency of action generation and keeps consistency with the pretraining paradigm. The total objective optimizes both components alongside the future-frame visual prediction loss in Eq.~\eqref{eq:loss}, where we set $\lambda_{\mathrm{tac}} = 1.0$ and $\lambda_{\mathrm{future}} = 0.5$.
\begin{equation}
\mathcal{L} = \mathcal{L}_{\mathrm{act}} + \lambda_{\mathrm{tac}}\,\mathcal{L}_{\mathrm{tac}} + \lambda_{\mathrm{future}}\,\mathcal{L}_{\mathrm{future}}
\label{eq:loss}
\end{equation}

\vspace{-2em}

\subsection{Training Recipe}
\label{sec:training_recipe}

\Ours is trained with a three-stage recipe that progressively transfers large-scale human visuomotor priors into tactile-reactive dexterous robot control.


\minisection{Large-scale Human Egocentric Pre-training}
Following EgoScale~\cite{zheng2026egoscale}, we pre-train the latent and action experts on 22,889 hours of egocentric human video. The latent expert learns visual and language representations from head-view observations, while the action expert is trained on retargeted human arm and hand motions represented in a unified action space. This stage provides broad semantic grounding and visuomotor priors for dexterous manipulation without tactile expert.

\minisection{Tactile Grounded Robot Mid-training}
Large-scale human pre-training provides broad visuomotor priors but limited grounding in robot-executable contact dynamics. We bridge this gap with 100 hours of teleoperated bimanual manipulation data with synchronized tactile signals, organized around diverse motor primitives and object interactions for compact coverage of contact-rich behaviors. This stage adapts the action expert to robot multiview observations and executable actions, while training the tactile expert to perform high-frequency denoising as a fine-grained refinement.

\minisection{Skill-Specific Post-training}
After tactile-grounded mid-training, \Ours already exhibits zero-shot contact-rich manipulation capabilities in Fig.~\ref{fig:ablation_midtrain}. For more complex or task-specific skills, we further fine-tune the model on approximately 100 task demonstrations, enabling it to adapt to specific task requirements while preserving the tactile-reactive behaviors acquired during mid-training.

\def\tabmainresults#1{
\begin{table*}[#1]
\centering
\tablestyle{2pt}{1.1}
\caption{\textbf{Comparison of \Ours and Baseline Methods Across 12 Tactile-reactive Manipulation Tasks.} Success rates (\%) are computed over 16 evaluation rollouts per task then across tasks.
}
\scriptsize
\setlength{\tabcolsep}{3.9 pt}

\resizebox{\textwidth}{!}{%
\begin{tabular}{l|cccccccccccc|c}
\toprule
Method
& \makecell[c]{Flip\\Page}
& \makecell[c]{Transfer\\Egg}
& \makecell[c]{Wipe\\Plate}
& \makecell[c]{Apply\\Paste}
& \makecell[c]{Split\\Cup}
& \makecell[c]{Sort\\Mahjong}
& \makecell[c]{Open\\Lock}
& \makecell[c]{Refill\\Tablet}
& \makecell[c]{Acid-Base\\Neut.}
& \makecell[c]{Extract\\Card}
& \makecell[c]{Deal\\Poker}
& \makecell[c]{Screw\\Bulb}
& Avg. \\
\midrule
ViTacFormer~\cite{heng2025vitacformer} & 9 & 0 & 4 & 1 & 4 & 7 & 0 & 0 & 0 & 2 & 2 & 1 & 3 \\
RDP~\cite{xue2025reactive} & 12 & 8 & 18 & 2 & 6 & 9 & 2 & 0 & 0 & 1 & 2 & 7 & 6 \\
Tactile-VLA~\cite{Huang2025TactileVLA} & 38 & 14 & 24 & 0 & 21 & 27 & 8 & 0 & 9 & 4 & 11 & 18 & 15 \\
EgoScale~\cite{zheng2026egoscale} & 68 & 44 & 34 & 38 & 33 & 36 & 19 & 12 & 43 & 41 & 28 & 18 & 35 \\
$\pi_{0.5}$~\cite{intelligence2025pi05visionlanguageactionmodelopenworld} & 36 & 17 & 28 & 13 & 18 & 32 & 5 & 1 & 24 & 8 & 9 & 11 & 17 \\
$\pi_{0.5}$ + tactile & 8 & 9 & 27 & 2 & 4 & 14 & 2 & 0 & 7 & 3 & 0 & 0 & 6 \\
\hline
\rowcolor{gray!10} Ours & \textbf{96} & \textbf{75} & \textbf{69} & \textbf{66} & \textbf{78} & \textbf{65} & \textbf{47} & \textbf{41} & \textbf{76} & \textbf{70} & \textbf{57} & \textbf{35} & \textbf{65} \\
\end{tabular}%
}

\label{tab:main_result}
\end{table*}
}

\section{Experiments}
\label{sec:exp}


\subsection{Experiment Setup}
\minisection{Robot Platform and Action Space}
As shown in Fig.~\ref{fig:robotsys} and App.~\ref{apdx:real-setup}, all real-world experiments use a fixed-base bimanual Dexmate Vega-1 robot with two 22-DoF Sharpa Wave dexterous hands. The policy observes RGB images from a ZED head camera and two monocular wrist cameras, along with per-finger tactile force vectors and deformation maps. Actions use relative end-effector delta control for the bimanual arms, and absolute joint control for the fingers.

\minisection{Baselines}
We compare T-Rex with 6 baselines: (1) ViTacFormer, an ACT-style visuo-tactile dexterous imitation policy with cross-attention fusion and future tactile prediction; 
(2) RDP, a slow-fast visual-tactile diffusion policy that performs high-frequency tactile-reactive action refinement for contact-rich manipulation;
(3) Tactile-VLA, a tactile-aware VLA model that integrates tactile sensing for contact-rich manipulation reasoning;
(4) $\pi_{0.5}$ and 
(5) EgoScale, a large-scale pretrained VLA foundation models fine-tuned on our task-specific post-training data; and 
(6) $\pi_{0.5}$ + tactile, which additionally conditions $\pi_{0.5}$ on tactile force signals and robot state. All methods use the same robot setup, action space, and evaluation protocol, with implementation details are provided in App.~\ref{apdx:baselines}.

\minisection{Evaluation Protocol and Metrics}
We evaluate all methods on the 12 tactile-reactive tasks defined in App.~\ref{apdx:tactile-reative_tasks}. For each task, we test for 16 trials, with object positions and rotations randomized across trials. We report average task success rate, using progress-based rubrics for multi-stage tasks to capture partial completion. Results are averaged across trials and then across tasks.


\subsection{Main Results}
\label{sec:main_results}

We evaluate \Ours against representative dexterous manipulation and VLA baselines on 12 tactile-reactive manipulation tasks requiring delicate force control, deformable object handling, and joint force-deformation reasoning. As shown in Tab.~\ref{tab:main_result}, T-Rex achieves the highest average success rate across all task categories, outperforming the strongest baseline by more than \textbf{30\%}.

\tabmainresults{!h}

Two key observations emerge. First, large-scale pre-training is essential for dexterous manipulation. Small policies trained from scratch on only 100 task-specific demonstrations, such as VitacFormer and RDP, consistently underperform across all tasks. Among the baselines, EgoScale achieves the strongest performance due to its large-scale egocentric pre-training with hand-pose supervision, substantially outperforming methods pre-trained only on task-specific real robot data like $\pi_{0.5}$ and Tactile-VLA. Second, tactile feedback is critical for contact-rich manipulation. Pretrained VLA models like EgoScale still fails at precise contact adjustment and force-sensitive behaviors. By combining large-scale pre-training with tactile-grounded mid-training and tactile-reactive control, \Ours achieves the strongest overall performance. We also observe that naively conditioning pretrained VLA models on tactile signals as $\pi_{0.5}$ + tactile can degrade performance, highlighting the importance of effective tactile integration. Further failure case analysis are provided in App.~\ref{apdx:fail_case}.

\begin{table}[t]
\centering
\tablestyle{7.2 pt}{1}
\caption{\textbf{Ablation Studies on Tactile Modality and Architectural Design.} Results are reported on six representative tactile-reactive manipulation tasks, with the final column showing the average.}
\begin{tabular}{@{} l | c c c c c c | l @{}}
\toprule
\textbf{Configuration} 
& \begin{tabular}[c]{@{}c@{}}Flip\\ Page\end{tabular} 
& \begin{tabular}[c]{@{}c@{}}Apply\\ Toothpaste\end{tabular} 
& \begin{tabular}[c]{@{}c@{}}Split\\ Cup\end{tabular} 
& \begin{tabular}[c]{@{}c@{}}Open\\ Lock\end{tabular} 
& \begin{tabular}[c]{@{}c@{}}Extract\\ Card\end{tabular} 
& \begin{tabular}[c]{@{}c@{}}Screw\\ Lightbulb\end{tabular} 
& \textbf{Average} \\ 
\midrule

\textbf{Full Model (Ours)} & \textbf{96} & \textbf{66} & \textbf{78} & \textbf{47} & \textbf{70} & \textbf{35} & \textbf{65} \\

\midrule
\rowcolor{gray!10} \multicolumn{8}{l}{\textit{Tactile Modality Ablation}} \\ 
w/o Tactile      & 76 & 39 & 58 & 23 & 34 & 20 & 42 {\color{DeepGreen}(-23\%)} \\
MLP Force + Deform       & 89 & 58 & 72 & 44 & 58 & 29 & 58 {\color{DeepGreen}(-7\%)} \\
Deform        & 82 & 57 & 71 & 36 & 55 & 25 & 54 {\color{DeepGreen}(-11\%)} \\
MLP Force + VQVAE Force      & 92 & 63 & 65 & 38 & 67 & 28 & 59 {\color{DeepGreen}(-6\%)} \\

\midrule
\rowcolor{gray!10} \multicolumn{8}{l}{\textit{Architecture Design}} \\ 
w/o Async        & 92 & 61 & 73 & 45 & 59 & 30 & 60 {\color{DeepGreen}(-5\%)} \\
\bottomrule
\end{tabular}
\label{tab:combined_ablation}
\end{table}

\subsection{Ablation Studies}
\label{sec:ablations}


\minisection{Impact of Dynamic Tactile Encoding and Representations}
We study the contribution of tactile information and tactile representations through a series of ablations. Specifically, we compare removing all tactile inputs (\textit{w/o tactile}), removing the proposed VQ-VAE force encoder while retaining the lightweight MLP and deformation signals (\textit{MLP Force + Deform}), using only deformation signals (\textit{Deform}), and using only force signals (\textit{MLP Force + VQ-VAE Force}). As shown in the upper section of Tab.~\ref{tab:combined_ablation}, these experiments evaluate the importance of tactile feedback, spatial deformation sensing, and the proposed temporal force encoding for tactile-reactive manipulation.

\begin{figure}[!h]
    \centering
    \includegraphics[width=\linewidth]{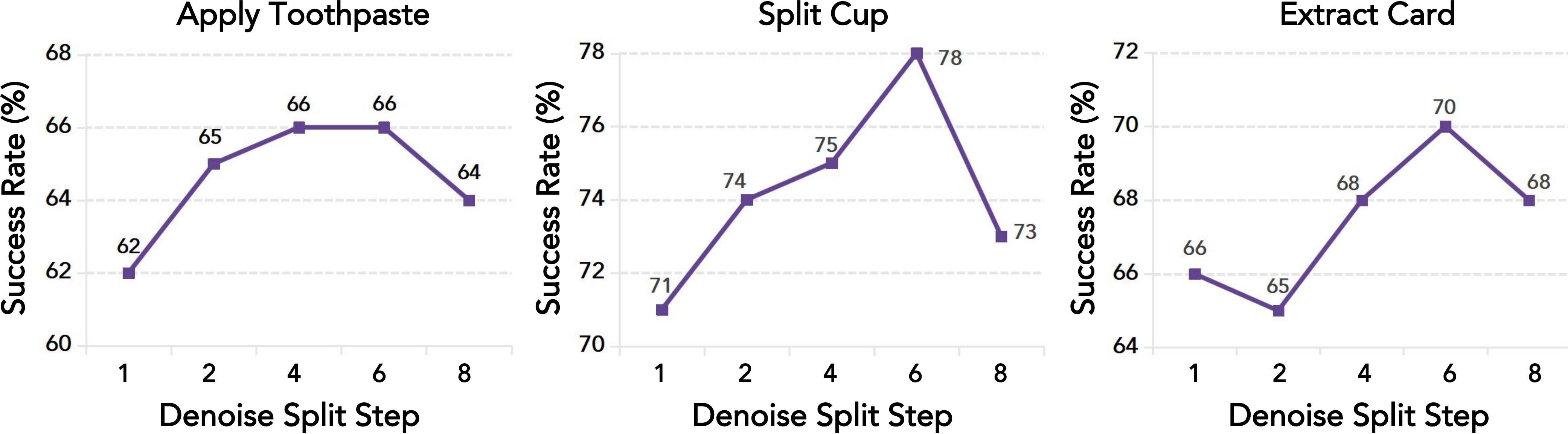}
    \caption{\textbf{Ablation Studies on Cascaded Denoising Split Steps $K_{\mathrm{slow}}$.} We show the success rate curve of different split steps.}
    \label{fig:ablation_split_step}
\end{figure}

\begin{figure}[!h]
    \centering
    \includegraphics[width=\linewidth]{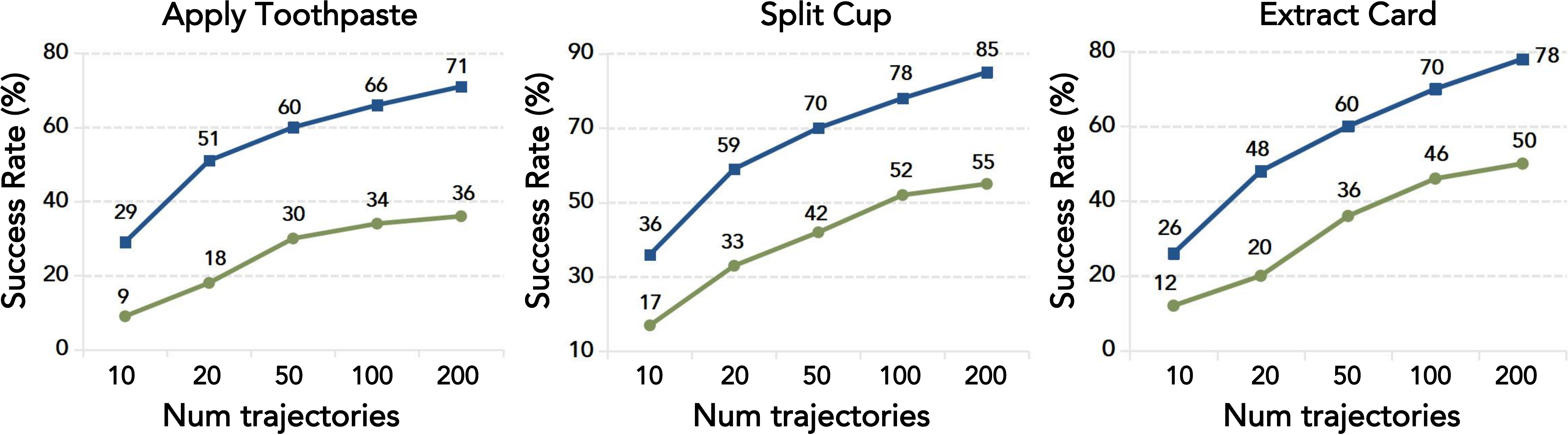}
    \caption{\textbf{Data Efficiency of T-Rex.} We show the success rate curve of different numbers of demonstrations. \textbf{\color{DeepBlue}Blue:} with our tactile-grounded T-Rex mid-training data; \textbf{\color{DeepGreen}Green:} without mid-training.}
    \label{fig:data_eff}
    \vspace{-0.1cm}
\end{figure}


\minisection{Impact of Asynchronous Tactile-Reactive Cascaded Flow Matching}
We compare the proposed asynchronous tactile refinement against a synchronous baseline. As shown in Tab.~\ref{tab:combined_ablation}, asynchronous refinement consistently improves performance, validating the benefit of decoupling low-frequency visuomotor planning from high-frequency tactile control. We further vary the denoising split step $\tau_{\mathrm{split}}$. As shown in Fig.~\ref{fig:ablation_split_step}, an intermediate split achieves the best performance. When $\tau_{\mathrm{split}}$ is too small, the action expert provides insufficient visuomotor priors for downstream refinement; when $\tau_{\mathrm{split}}$ is too large, the tactile expert has limited capacity to incorporate tactile feedback.

\begin{figure}[!h]
    \centering
    \includegraphics[width=\linewidth]{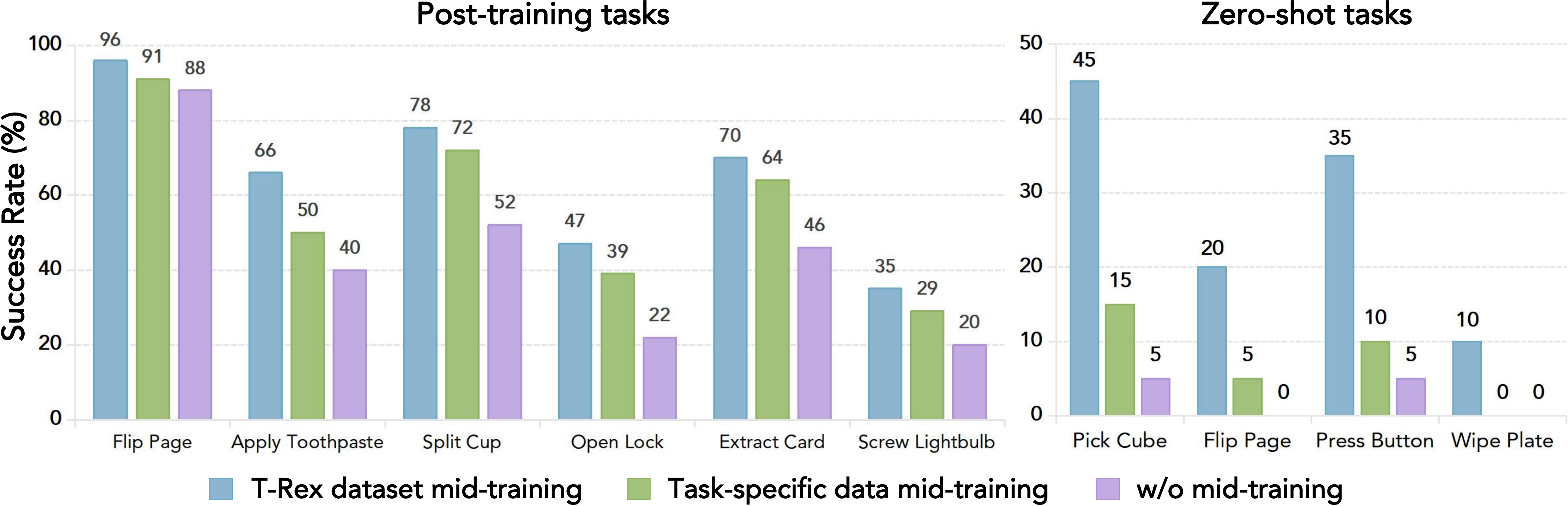}
    \caption{\textbf{Ablation Studies on Mid-training Datasets.} We select 6 representative tasks for post training evaluation and 4 easier tasks for zero-shot evaluation, including motor primitives of \textit{pick, slide, press and wipe} in T-Rex dataset.}
    \label{fig:ablation_midtrain}
    \vspace{-0.1cm}
\end{figure}

\minisection{Efficiency of Tactile-Grounded \Ours Dataset}
We compare the proposed 100-hour tactile-grounded \Ours Dataset with a 100-hour task-specific dataset collected from 11 tasks, ensuring a matched data budget. As shown in Fig.~\ref{fig:ablation_midtrain}, the proposed dataset achieves stronger generalization and zero-shot transfer. We further vary the number of post-training demonstrations from 10 to 200. As shown in Fig.~\ref{fig:data_eff}, tactile-grounded mid-training substantially improves performance in the low-data regime, reducing the amount of downstream data required for contact-rich dexterous manipulation.

\begin{table}[!h]
\centering
\vspace{-0.2cm}
\caption{\textbf{Effectiveness of the Training Recipe of \Ours.} We selected six representative tasks and report the success rates and compare the success rate (\%) on different training recipes.}
\tablestyle{7pt}{1}
\begin{tabular}{cc|ccccccc}
\toprule
Pre-training & Mid-training & \begin{tabular}[c]{@{}c@{}}Flip\\ Page\end{tabular} & \begin{tabular}[c]{@{}c@{}}Apply\\ Toothpaste\end{tabular} & \begin{tabular}[c]{@{}c@{}}Split\\ Cup\end{tabular} & \begin{tabular}[c]{@{}c@{}}Open\\ Lock\end{tabular} & \begin{tabular}[c]{@{}c@{}}Extract\\ Card\end{tabular} & \begin{tabular}[c]{@{}c@{}}Screw\\ Lightbulb\end{tabular} & Average \\ \midrule 

\multirow{2}{*}{\coloredcross{Firebrick}}
& \coloredcross{Firebrick}
& 46 & 16  & 20 & 6  & 14 & 5  & 18 \\ 
& \coloredcheckmark{ForestGreen}
& 75 & 34 & 45 & 10 & 32 & 9 & 34 \\ 
\multirow{2}{*}{\coloredcheckmark{ForestGreen}}
& \coloredcross{Firebrick}
& 88 & 40 & 52 & 22 & 46 & 20 & 45 \\
& \coloredcheckmark{ForestGreen}
& 96 & 66 & 78 & 47 & 70 & 35 & 65 \\ \bottomrule
\end{tabular}
\label{tab:ablation_recipe}
\vspace{-0.1cm}
\end{table}

\minisection{Effectiveness of the Training Recipe}
Finally, we validate the proposed three-stage training recipe by ablating large-scale human egocentric pretraining and tactile-grounded mid-training. Specifically, we compare variants with and without human pretraining, and with and without tactile-grounded mid-training, on six robot tasks from our benchmark. This study isolates the role of each stage: human pretraining provides broad semantic grounding and coarse visuomotor priors, while tactile-grounded mid-training bridges these priors to robot-executable contact-rich control. Results in Tab.~\ref{tab:ablation_recipe} show both stages contribute to performance, with the full recipe achieving the best results.

\section{Conclusion}
\label{sec:conc}
We enable foundational manipulation policies to achieve scalable, tactile-reactive dexterous control. 
We introduce \Ours, a Mixture-of-Transformer-Experts (MoT) model utilizing asynchronous tactile refinement and a dynamic tactile VAE encoding. 
Our framework leverages general human video pre-training, followed by mid-training on our newly contributed, open-source 100-hour tactile-synchronized dexterous manipulation dataset. 
Post-trained and evaluated across 12 real-world tactile-reactive tasks, \Ours outperforms existing dexterous and tactile-aware VLA baselines by an average success rate of 30\% and significantly improve data efficiency. 

\section{Limitation and Future Work}
\label{sec:limitation}

While \Ours demonstrates strong performance and data efficiency, it highlights several avenues for future research. First, for long-horizon tasks with precise contact coordination and tight tolerances where teleoperation is difficult, future work could integrate reinforcement learning or online interaction-based refinement. Second, tactile-reactive manipulation remains bottlenecked by hardware, including sensor distortion, calibration drift across devices, and the absence of dense palm sensing for whole-hand manipulation. Future work may explore unified representations across heterogeneous tactile sensors and richer, whole-hand tactile hardware.

\acknowledgments{
We thank Sharpa for providing maintenance updates for their equipment. We also thank Yusuke Kato from Panasonic for his contributions to the collection of part of the T-Rex dataset. UC Berkeley authors were supported in part by the Berkeley Artificial Intelligence Research Humanoid Intelligence Center (BAIR HIC). Sapienza University acknowledges funding from Panasonic and from the Sapienza grant RG123188B3EF6A80 (CENTS). We thank Alessio Sampieri and Luca Franco (ItalAI S.r.l.) for fruitful discussions.}

\clearpage

\bibliography{references}  

\clearpage
\appendix
{\Large\bfseries Appendix\par}
\addcontentsline{toc}{section}{Appendix}
\vspace{0.3cm}


In this appendix, we first present the model architecture and training hyperparameters in App.~\ref{apdx:model-training}. We then provide additional details of the proposed asynchronous tactile-reactive cascaded denoising framework in App.~\ref{apdx:add_method}, followed by implementation details of alternative tactile encoders used to validate the proposed spatio-temporal tactile representation in App.~\ref{apdx:tacile_encoder}. Next, we describe the real-world experimental setup in App.~\ref{apdx:real-setup}. App.~\ref{apdx:baselines} and App.~\ref{apdx:tactile-reative_tasks} provide implementation details of the baselines and benchmark tasks, including evaluation protocols and scoring criteria. We further present the construction and composition of the T-Rex dataset in App.~\ref{apdx:trex_dataset}. Finally, App.~\ref{apdx:fail_case} presents representative failure cases and discusses future directions for tactile-reactive dexterous manipulation.


\section{Model and Training Details}
\label{apdx:model-training}

Detailed model architectures and training hyperparameters for T-Rex are summarized in Tab.~\ref{tab:training_details}.

\begin{table}[htbp]
    \centering
    \caption{Model and Training Configurations for T-Rex.}
    \label{tab:training_details}
    \tablestyle{16 pt}{1.1}
    \begin{tabular}{l c}
        \toprule
        
        \multicolumn{2}{c}{\textit{Latent Expert}} \\
        \midrule
        VLM Backbone & Qwen3VL-2B \\
        Hidden Feature Dimension & 2048 \\
        Transformer Layers & 28 \\
        Max Sequence Length & 2048 \\
        Parameter size & 1.41B \\
        Attention Implementation & Flash Attention 2 \\
        
        \midrule
        \multicolumn{2}{c}{\textit{Action Expert (Flow Matching)}} \\
        \midrule
        VLM Backbone & Qwen3VL-2B \\
        Action Dimension & 62 \\
        Action Chunk & 16 \\
        Training Timestep Sampling & $\mathrm{Beta}(1.5, 1.0)$ \\
        Num Inference Timesteps & 6 \\
        Parameter size & 1.41B \\

        \midrule
        \multicolumn{2}{c}{\textit{Tactile Expert (Flow Matching)}} \\
        \midrule
        Action Dimension & 62 \\
        Action Chunk & 16 \\
        FFN Intermediate Size & 1536 \\
        Training Timestep Sampling & $\mathrm{Beta}(1.5, 1.0)$ \\
        Num Inference Timesteps & 4 \\
        Parameter size & 0.62B \\
        
        \midrule
        \multicolumn{2}{c}{\textit{Training Configurations (SFT)}} \\
        \midrule
        Optimizer & AdamW \\
        Peak Learning Rate & $1 \times 10^{-4}$ \\
        Min Learning Rate & 0 \\
        LR Scheduler & Cosine with $\min$ LR \\
        Weight Decay & 0 \\
        Warmup Ratio & 0 \\
        Gradient Clipping & 1.0 \\
        GPU Type & NVIDIA H100 \\
        Number of GPUs & 24 \\
        Deepspeed Zero Stage & 1 \\
        Per Device Batch Size & 16 \\
        Gradient Accumulation Steps & 1 \\
        Mixed Precision Training & bf16 \\
        \bottomrule
    \end{tabular}
\end{table}

\section{Additional Details for Asynchronous Cascaded Denoising} 
\label{apdx:add_method}

Building upon the macroscopic formulation introduced in Section~\ref{sec:cascaded_flow}, we provide the exact optimization objectives, conditioning contexts, and runtime implementation details essential for the asynchronous tactile-reactive cascaded flow matching. The complete inference procedure is formalized in Algorithm~\ref{alg:cascaded_inference}.

\minisection{Explicit Conditioning and Training Objectives}
During training, the two experts regress the shared velocity target $v^\star$ but are conditioned on distinctly different contexts to enforce their respective specialized roles. The action expert is conditioned exclusively on the multimodal latent context $\mathbf{c}^{\mathrm{vl}}$ (comprising head/wrist camera features, language prompts, and future-prediction tokens). Its objective is given by:
\begin{equation}
\mathcal{L}_{\mathrm{act}} = \mathbb{E}\bigl\lVert f_\theta^{\mathrm{act}}(\mathbf{x}_{\tau_{\mathrm{act}}}, \tau_{\mathrm{act}}; \mathbf{c}^{\mathrm{vl}}) - v^\star \bigr\rVert_2^2
\end{equation}
Conversely, the tactile expert operates completely independent of the raw visual observations. Instead, it is conditioned on the high-frequency tactile tokens $\mathbf{c}^{\mathrm{tac}}$ and the detached intermediate state from the slow stream. Specifically, we execute the slow tick under \texttt{torch.no\_grad} to obtain the key-value cache $\mathrm{KV}_{\tau_{\mathrm{split}}}$. The tactile expert's objective is defined as:
\begin{equation}
\mathcal{L}_{\mathrm{tac}} = \mathbb{E}\bigl\lVert f_\theta^{\mathrm{tac}}(\mathbf{x}_{\tau_{\mathrm{tac}}}, \tau_{\mathrm{tac}}; \mathbf{c}^{\mathrm{tac}}, \mathrm{KV}_{\tau_{\mathrm{split}}}) - v^\star \bigr\rVert_2^2
\end{equation}
The total objective jointly optimizes both components alongside the future-frame visual prediction loss (Sec. \ref{sec:model_architecture}):
\begin{equation}
\mathcal{L} = \mathcal{L}_{\mathrm{act}} + \lambda_{\mathrm{tac}}\mathcal{L}_{\mathrm{tac}} + \lambda_{\mathrm{future}}\mathcal{L}_{\mathrm{future}}, \qquad \text{where } \lambda_{\mathrm{tac}} = 1.0, \;\; \lambda_{\mathrm{future}} = 0.5.
\end{equation}

\minisection{KV Cache Composition and Delay Augmentation}
The refreshed cache passed to the tactile expert is formally composed as $\mathrm{KV}_{\tau_{\mathrm{split}}} = \bigl[\mathrm{KV}^{\mathrm{lat}} \big| \mathrm{KV}^{\mathrm{act}}_{\tau_{\mathrm{split}}}\bigr]$, which contains both the visual-language keys/values and the action positions re-encoded at time $\tau_{\mathrm{split}}$. This re-encoding ensures the tactile expert attends to a coherent, partially-denoised contextual manifold rather than the initial noise-time encoding. 

Furthermore, because the fast ticks in deployment run asynchronously at intra-chunk offsets, there is an inherent temporal staleness between the frozen visual cache and the real-time tactile stream. To prevent the policy from overfitting to perfectly synchronized modalities during mid-training, we introduce a \textit{delay augmentation}. We draw a discrete delay $\delta \sim \mathrm{Uniform}\{0, 4, 8, 12\}$ to randomly shift the frame indices used for extracting $\mathbf{c}^{\mathrm{tac}}$ relative to those used for $\mathbf{c}^{\mathrm{vl}}$, strictly matching the deployment-time staleness distribution.

\minisection{Computational Amortization and Runtime Synchronization}
The cascaded design yields substantial computational savings during deployment. Crucially, the visual tower, the latent expert, and the action expert do not re-execute during a fast tick. The per-control-step computational cost is therefore dominated exclusively by the $K_{\mathrm{fast}}$ Euler steps of the lightweight tactile expert (which utilizes a reduced FFN intermediate size, as detailed in Tab~\ref{tab:training_details}).

To ensure thread safety between the parallel asynchronous streams on the real robot, the deployment runtime utilizes a single-threaded request socket combined with an explicit execution lock. As detailed in Algorithm~\ref{alg:cascaded_inference}, this mechanism serializes the two experts, guaranteeing that no high-frequency fast tick initiates until any in-flight slow tick has fully committed its $\mathrm{KV}_{\tau_{\mathrm{split}}}$ cache and intermediate boundary state $\hat{\mathbf{x}}_{\tau_{\mathrm{split}}}$ to the shared memory space.

\newcommand{\bx}{\mathbf{x}}
\newcommand{\bA}{\mathbf{A}}
\newcommand{\bc}{\mathbf{c}}
\newcommand{\tausplit}{\tau_{\mathrm{split}}}

\begin{algorithm}[t]
\caption{Asynchronous Tactile-Reactive Cascaded Flow Matching Inference}
\label{alg:cascaded_inference}
\begin{algorithmic}[1]

\Require 
    Pre-trained experts $f_\theta^{\mathrm{act}}$ and $f_\theta^{\mathrm{tac}}$; 
    Total flow steps $N$, slow segment steps $K_{\mathrm{slow}}$ ($K_{\mathrm{fast}} = N - K_{\mathrm{slow}}$); 
    Step size $\Delta\tau = -1/N$; Boundary threshold $\tausplit = 1 - K_{\mathrm{slow}}/N$.
\Ensure 
    Executed actions $\bA_{t:t+T_a}$ at corresponding execution offsets.

\Statex
\State \textbf{Shared Memory:} Intermediate state $\hat{\bx}_{\tausplit}$, KV Cache $\mathrm{KV}_{\tausplit}$, Execution Lock $\text{lock}$
\Statex

\setstretch{1.1}
\parbox{\linewidth}{
\begin{minipage}[t]{0.47\linewidth}
    \Procedure{Slow-Stream Loop (LowFreq)}{}
        \For{each action chunk window $T_a$}
            \State Get vision-language context $\bc^{\mathrm{vl}}$
            \State Sample initial noise $\bx_1 \sim \mathcal{N}(\mathbf{0}, \mathbf{I})$
            \Statex \hfill $\triangleright$ \textit{Upper Segment Integration}
            \For{$k = 1$ \textbf{to} $K_{\mathrm{slow}}$}
                \State $\tau \gets 1 - (k-1)/N$
                \State $v \gets f_\theta^{\mathrm{act}}(\bx_\tau, \tau;\, \bc^{\mathrm{vl}})$
                \State $\bx_{\tau + \Delta\tau} \gets \bx_\tau + \Delta\tau \cdot v$
            \EndFor
            \Statex
            \State \textbf{acquire} $\text{lock}$
            \State $\hat{\bx}_{\tausplit} \gets \bx_{\tausplit}$
            \State Refresh and re-encode position cache:
            \Statex \hfill $\mathrm{KV}_{\tausplit} \gets \bigl[\,\mathrm{KV}^{\mathrm{lat}}\;\big|\;\mathrm{KV}^{\mathrm{act}}_{\tausplit}\,\bigr]$
            \State \textbf{release} $\text{lock}$
        \EndFor
    \EndProcedure
\end{minipage}
\hfill
\begin{minipage}[t]{0.47\linewidth}
    \Procedure{Fast-Stream Loop (HighFreq)}{}
        \For{offsets $\delta \in \{0, 4, 8, 12\}$ inside window}
            \State Sample real-time tactile stream $\bc^{\mathrm{tac}}$
            \Statex
            \State \textbf{acquire} $\text{lock}$
            \State Clone context: $\mathbf{kv} \gets \text{clone}(\mathrm{KV}_{\tausplit})$
            \State $\bx \gets \hat{\bx}_{\tausplit}$
            \State \textbf{release} $\text{lock}$
            \Statex \hfill $\triangleright$ \textit{Terminal Segment Integration}
            \For{$k = 1$ \textbf{to} $K_{\mathrm{fast}}$}
                \State $\tau \gets \tausplit - (k-1)/N$
                \State $v \gets f_\theta^{\mathrm{tac}}(\bx, \tau;\, \bc^{\mathrm{tac}}, \mathbf{kv})$
                \State $\bx \gets \bx + \Delta\tau \cdot v$
            \EndFor
            \Statex
            \State $\hat{\bA}_{t+\delta:t+\delta+T_a} \gets \bx$
            \State \textbf{Execute} updated action chunk
        \EndFor
    \EndProcedure
\end{minipage}
}

\end{algorithmic}
\end{algorithm}

\section{Implementation Details for Spacial-Temporal Tactile Encoder} 
\label{apdx:tacile_encoder}
\minisection{VQ-VAE Dynamic Force Encoder} 
To robustly process high-frequency tactile observations and mitigate inherent sensor drift, continuous multi-finger force sequences are discretized into a compact token space using a Vector-Quantized Variational Autoencoder (VQ-VAE)~\cite{oord2018neuraldiscreterepresentationlearning}. 

For each fingertip, raw six-dimensional force vectors are collected over a short temporal window of $T=16$ frames. The VQ-VAE encoder consists of a 1D temporal convolutional network that hierarchically downsamples the temporal dimension via two strided blocks, followed by temporal mean-pooling to produce a $256$-dimensional continuous embedding. This embedding is subsequently mapped by a vector quantizer to its nearest neighbor within a learned codebook of size $K=64$. The codebook parameters are updated via an Exponential Moving Average (EMA), where underutilized codebook entries are periodically re-seeded from current batch activations to prevent codebook collapse.

Meanwhile, a symmetric decoder is employed to reconstruct the original force sequence from the quantized tokens. To prevent the codebook from collapsing onto dominant non-contact states, the network is optimized via a magnitude-weighted Mean-Squared Error (MSE) loss, which assigns higher optimization penalties to frames experiencing high-force contacts. To maintain parameter efficiency and cross-digit scalability, convolutional weights are shared across all five fingers, with distinct learned finger-identity embeddings injected prior to encoding. This architecture compresses noisy, high-dimensional tactile inputs into one discrete, drift-robust token per finger per hand, forming a structured tactile vocabulary that is subsequently consumed by the fast tactile expert alongside spatial deformation maps.

\minisection{Tactile Deformation Encoder}
Complementing the temporal force profiles, each fingertip simultaneously provides a dense, single-channel spatial deformation map $\mathbf{d}_t$ representing the local skin displacement field. These maps capture rich, high-frequency contact geometry, such as edges, slip, and shear patterns that are inherently lost in low-dimensional force vectors. 

To process these maps, we employ a lightweight convolutional network adapted from a ResNet-18 backbone~\cite{he2015deepresiduallearningimage}. The standard input stem is modified to ingest a single-channel input, and only the first three residual stages are retained. Each stage is appended with a $3\times3$ convolutional layer that re-projects the intermediate feature maps to a fixed width of $128$ channels. The resulting spatial feature tensor is flattened and linearly projected into the policy's token space. To supply a stable, geometry-aware contact representation without expanding the trainable parameter footprint of the policy network, this encoder is pre-trained within a self-supervised convolutional autoencoder framework and subsequently frozen during policy learning~\cite{huang2026spatiallyanchoredtactileawareness}. During fast-stream inference, these per-fingertip deformation embeddings are concatenated with the quantized force tokens, yielding the complete, unified tactile observation consumed by the tactile expert.

\section{Real-World Setup and Teleoperation Stack}
\label{apdx:real-setup}

\begin{figure}[h]
    \centering
    \includegraphics[width=\linewidth]{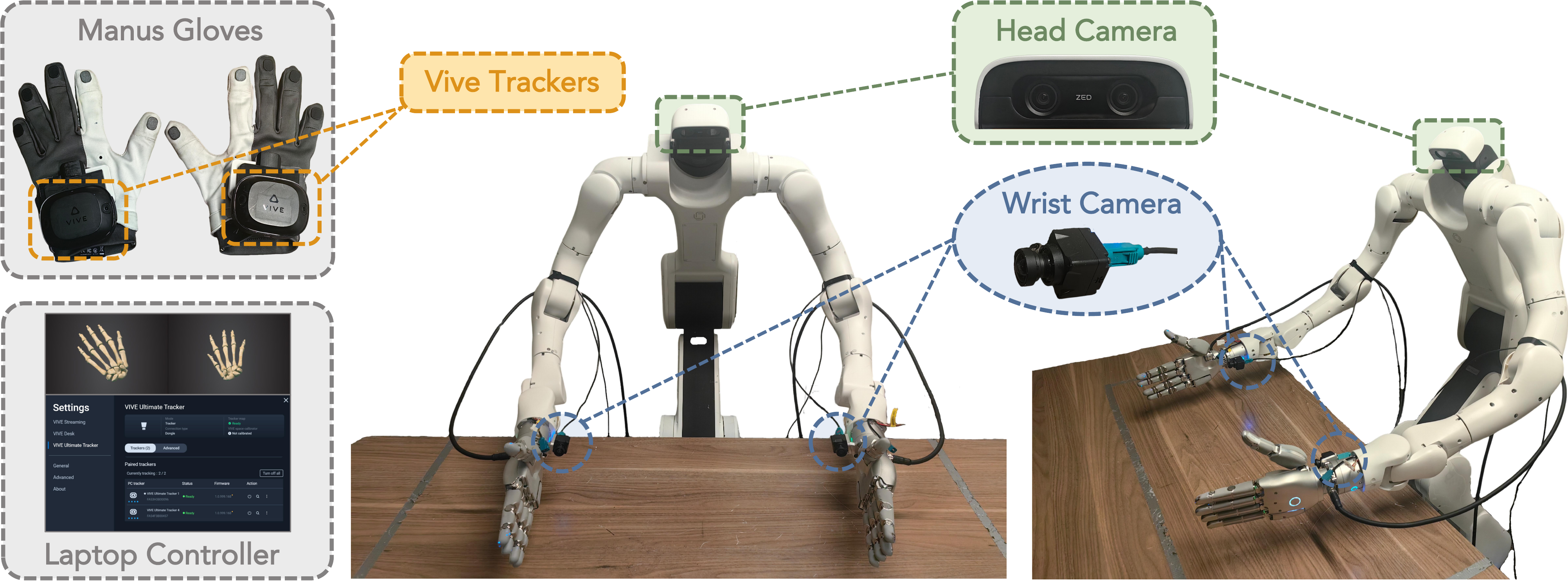}
    \caption{Robot system setup on the Dexmate Vega-1 bimanual robot and the Sharpa Wave dexterous hands. Two ZED X One S (wide view) cameras are mounted at the wrists, and one ZED X Mini camera is mounted on the head. For teleoperation we use Manus gloves to retrieve hand target gesture and VIVE trackers for wrist target pose. }
    \label{fig:robotsys}
\end{figure}

We conduct data collection and policy rollout on a Dexmate Vega-1 bimanual robot equipped with two Sharpa Wave dexterous hands. This section describes the hardware, perception system, and teleoperation interface used in our experiments. An overview of the system is shown in Fig.~\ref{fig:robotsys}.

\minisection{Robot Hardware and Control}
The Dexmate Vega-1 is a dual-arm mobile robot with 7 actuated joints per arm. In our setup, we keep the wheels, torso, and head joints fixed, and actuate only the 14 arm joints. To control the robot using relative end-effector pose commands, we use differential inverse kinematics through Pink~\cite{pink}. The resulting joint-space commands are passed through a low-pass filter before being sent to the manufacturer's low-level cascade PID controller. During policy rollout, a \Ours policy inference thread runs concurrently with a high-frequency low-level control thread operating at 300~Hz. The policy outputs action chunks, which asynchronously update the targets tracked by the low-level controller.

\minisection{Perception System}
The Dexmate Vega-1 includes a ZED X Mini stereo camera mounted on the head. We use the left monocular RGB stream from this camera. In addition, we mount two ZED X One S monocular RGB cameras (wide-view variant) on the robot wrists to capture viewpoints that may be occluded from the head camera. The camera poses are adjusted so that the head camera observes the full reachable workspace in front of the robot, while the wrist cameras maintain clear views of the fingers without significant occlusion from the palms. All three RGB streams are captured at a resolution of \(640 \times 360\). In addition to visual observations, each robot hand contains five fingertip tactile sensors. For each tactile sensor, we record and use the estimated deformation depth and the 6-axis net wrench.

\minisection{Teleoperation}
For real-world data collection, we use a human teleoperation system based on Manus gloves and VIVE trackers. The two VIVE trackers provide \(SE(3)\) wrist poses, which are passed through the same control pipeline used during policy rollout. The Manus gloves provide fingertip positions relative to the hand bases. These positions are retargeted to the Sharpa Wave robot hands using a manufacturer-provided differential inverse kinematics package based on Pinocchio~\cite{pinocchio} and CasADi~\cite{casadi}. As in policy rollout, teleoperation uses a high-level thread and a high-frequency low-level control thread. The high-level thread runs at 30~Hz, reads target pose information from the Manus gloves and VIVE trackers, retargets the commands to robot joint space, records video and proprioceptive observations, and asynchronously updates the 300~Hz low-level control thread.

\section{Implementation Details of Baselines} \label{apdx:baselines}

In the Sec.~\ref{sec:main_results} of our main paper, we compare \Ours with 6 baselines across 12 tasks, here we provide the implementation details of reproduce of the 6 baselines.

\textbf{ViTacFormer}~\citep{heng2025vitacformer} is an ACT-style visuo-tactile imitation learning policy that learns cross-modal representations through visual-tactile fusion and an auxiliary future tactile prediction objective. We follow the official implementation\footnote{\url{https://github.com/RoboVerseOrg/ViTacFormer}} and reproduce ViTacFormer as a task-specific baseline on our 12 contact-rich tasks. Specifically, we train separate ACT policies for each of the 12 T-Rex tasks using the same post-training setting as our method, with 100 demonstrations per task and 100 training epochs. Following the original design, we use 6D per-finger force vectors as tactile conditioning inputs and enable bimanual control for both arms. We use an ACT chunk size of 100, hidden dimension of 512, feedforward dimension of 3200, and KL weight of 10. The original implementation assumes 21-DoF dexterous hands with several mechanically coupled joints masked out during prediction. We adapt the policy to our 22-DoF Sharpa Wave hands and predict all finger joints directly without masking, enabled by the fully actuated hardware design. All models are trained with AdamW using a learning rate of $3\times10^{-4}$ and a global batch size of $16\times8$. The observation space, action space, and evaluation protocol are unified across all baselines.

\textbf{Reactive Diffusion Policy (RDP)}~\citep{xue2025reactive} is a slow-fast visuo-tactile imitation learning framework that combines a low-frequency latent diffusion policy with a high-frequency tactile-reactive controller for contact-rich manipulation. We follow the official implementation\footnote{\url{https://github.com/xiaoxiaoxh/reactive_diffusion_policy}} and reproduce RDP as a task-specific baseline on our 12 contact-rich tasks. Specifically, for each of the 12 T-Rex tasks, we separately train the Asymmetric Tokenizer (AT) and Latent Diffusion Policy (LDP) using the same post-training demonstrations as our method. For the AT stage, we train a tactile-conditioned action tokenizer for 100 epochs using a batch size of 64 and a learning rate of $1\times10^{-3}$. Following the original design, we use tactile force observations as high-frequency conditioning signals, where the tactile input consists of 6D force/torque vectors from all 10 fingers. For the LDP stage, we train the latent diffusion policy for 200 epochs initialized from the latest AT checkpoint. We use the original CNN-based diffusion architecture and slow-fast latent action formulation proposed in RDP. All models are trained separately per task using identical training splits and evaluation settings as other baselines.

\textbf{Tactile-VLA}~\citep{Huang2025TactileVLA} is a tactile-aware vision-language-action model that integrates tactile sensing into VLA policies for contact-rich manipulation through multimodal fusion and hybrid force-position control. Follow the paper, we reproduce Tactile-VLA as a task-specific baseline on our 12 contact-rich tasks. Since the original method uses GelSight tactile images as tactile inputs, we adapt the tactile encoder to instead use 6D force/torque vectors from all 10 fingers, matching the tactile observations available on our platform. Following the original design, we train separate Tactile-VLA policies for each of the 12 T-Rex tasks using the same post-training demonstrations as our method. All models are trained for 100 epochs on 8 GPUs using the Simple-MLP tactile encoder. We use a peak learning rate of $3\times10^{-4}$ with cosine decay to $3\times10^{-5}$ and linear warmup for the first 5300 steps. The observation space, action space, and evaluation protocol are unified across all baselines.

\textbf{EgoScale}~\citep{zheng2026egoscale} studies the scalability of large-scale egocentric human video pretraining for dexterous manipulation, showing that human action prediction improves predictably with data scale and transfers to high-DoF robotic hands. We reproduce this baseline using the GR00T N1.7 implementation\footnote{\url{https://github.com/Nvidia/Isaac-GR00T}} and initialize from the pretrained \texttt{nvidia/GR00T-N1.7-3B} checkpoint. For each of the 12 T-Rex tasks, we fine-tune a separate policy on the same task-specific demonstrations used in our post-training stage. Each policy is trained for 200 epochs with a global batch size of 32 on 8 GPUs. We use the relative end-effector actions for the bimanual arms, and 22-DoF joint actions for the Sharpa Wave hands. During fine-tuning, we apply state dropout with probability 0.2 and standard image color jitter augmentation. The observation space, action space, and evaluation protocol are kept the same as other baselines.

\textbf{$\pi_{0.5}$ and $\pi_{0.5}$ + tactile~\cite{black2024pi0}}
We reproduce $\pi_{0.5}$ using the official OpenPI codebase\footnote{\url{https://github.com/Physical-Intelligence/openpi}} and initialize all policies from the released $\pi_{0.5}$ pretrained checkpoint. For each of the 12 T-Rex tasks, we fine-tune separate policies using the same task-specific post-training demonstrations as our method. We adopt a bimanual joint-space control setup consisting of dual-arm $2\times 7$ joint control and 22-DoF dexterous hand joint control. 

We evaluate two variants: a visual-only $\pi_{0.5}$ baseline and a tactile-conditioned $\pi_{0.5}$ + tactile baseline. For the tactile version, we extend the original state input by concatenating single-step tactile observations consisting of 6D force/torque vectors from all 10 fingers. Following the official implementation, we use the $\pi_{0.5}$ action expert architecture with action horizon 16 and fine-tune using the provided cosine learning rate schedule with peak learning rate $5\times10^{-5}$. All models are trained on 8 GPUs with FSDP enabled and a global batch size of 16. The observation space, action space, and evaluation protocol are unified across all baselines.

\section{Evaluation Tasks} 
\label{apdx:tactile-reative_tasks}

We evaluate \Ours on 12 contact-rich dexterous manipulation tasks which capture various real world force-reactive and tactile-deformation situations.
%
Force-reactive tasks require the robot to precisely regulate contact forces during manipulation—such as grasping fragile objects, applying controlled pressure, or resisting slip. Success depends on tactile feedback to adjust grip force and avoid object damage or task failure.
Tactile-deformation sensitive tasks involve objects or mechanisms where deformation of the tactile sensor pad plays a key role—such as stacked cups, or mahjong tiles identified by surface texture. The robot must sense and respond to physical deformation that cannot be detected by vision alone.
Some tasks require both at the same time, often in longer sequences involving insertion, extraction, and bimanual handovers. They are the most challenging category among our 12 contact-rich tasks.
Each task is evaluated using one of two grading rubrics: an \textit{additive} rubric awards independent partial credit for each completed sub-step, while a \textit{progress-based} rubric assigns a single score reflecting how far the robot progressed along a predefined success hierarchy.


\begin{figure}[h]
    \centering
    \includegraphics[width=\linewidth]{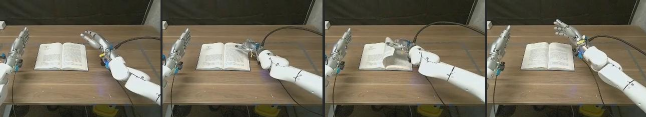}
    \caption{Key stages of Task~I: Flip Page.}
    \label{fig:flip_page}
\end{figure}

\noindent\textbf{Task I: Flip Page.} \textit{Text Instruction:}
``Turn a page of the book from right to left using your right index finger.'' The robot must lift
a single sheet from the right side of an open book, sweep it across the
spine, and smooth it down flat on the left side.

\textit{Grading rubric (additive):}
\begin{itemize}
    \item $+0.3$: \textbf{(a)} Successfully touched the book page with a single finger.
    \item $+0.3$: \textbf{(b)} Using the index finger turn the page up.
    \item $+0.4$: \textbf{(c)} Successfully flips exactly one page from right to left and smooths it flat.
\end{itemize}

\begin{figure}[h]
    \centering
    \includegraphics[width=\linewidth]{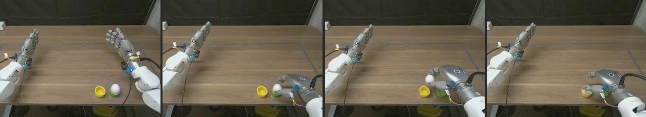}
    \caption{Key stages of Task~II: Transfer Egg.}
    \label{fig:transfer_egg}
\end{figure}

\noindent\textbf{Task II: Transfer Egg.} \textit{Text Instruction:}
``Using the right thumb and index finger, pick up the egg from the green egg tray and place it into the yellow egg tray.'' The
robot must grasp a fragile egg without cracking the shell from the green container, lift it off the surface,
transport it above the yellow container, and gently release it inside.

\textit{Grading rubric (additive):}
\begin{itemize}
    \item $+0.2$: \textbf{(a)} Approaches the egg and makes contact without knocking it off the table.
    \item $+0.3$: \textbf{(b)} Lifts the egg off the table without cracking it.
    \item $+0.2$: \textbf{(c)} Transports the egg above the yellow container.
    \item $+0.3$: \textbf{(d)} Releases the egg inside the container intact.
\end{itemize}

\begin{figure}[h]
    \centering
    \includegraphics[width=\linewidth]{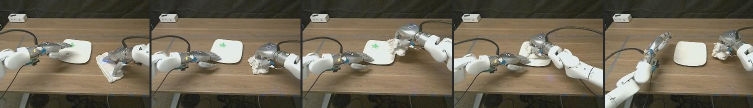}
    \caption{Key stages of Task~III: Wipe Plate.}
    \label{fig:wipe_plate}
\end{figure}

\noindent\textbf{Task III: Wipe Plate.} \textit{Text Instruction:}
``There is a white plate and a white cloth on the table; the white plate has colored stains on it. Use your right hand to pick up the cloth, hold the plate steady with your left hand, and then use the cloth to wipe away the stains.'' The
robot must grasp the cloth with the right hand, press down on the plate
with the left hand to hold it steady, bring the cloth into contact with
the plate surface, wipe the plate until the colored stains are fully
removed, and place the cloth back on the table while releasing the plate.

\textit{Grading rubric (additive):}
\begin{itemize}
    \item $+0.2$: \textbf{(a)} Right hand grasps the rag.
    \item $+0.1$: \textbf{(b)} Left hand presses down on the plate to hold it steady.
    \item $+0.2$: \textbf{(c)} Brings the rag into contact with the plate surface.
    \item $+0.4$: \textbf{(d)} Wipes the plate until the design is fully removed
          (no visible ink remaining).
    \item $+0.1$: \textbf{(e)} Places the rag back on the table and releases the plate.
\end{itemize}

\begin{figure}[h]
    \centering
    \includegraphics[width=\linewidth]{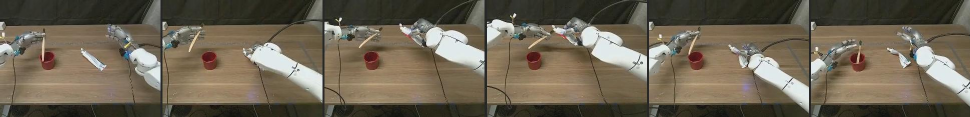}
    \caption{Key stages of Task~IV: Apply Toothpaste.}
    \label{fig:apply_toothpaste}
\end{figure}

\noindent\textbf{Task IV: Apply Toothpaste.} \textit{Text Instruction:}
``On the left side of the countertop sits a cup holding a toothbrush, while an open tube of toothpaste rests on the right. Pick up the toothbrush with your left hand and the toothpaste with your right, squeeze some toothpaste onto the brush, and then set the tube back down.'' The
robot must grasp a toothbrush in one hand and a toothpaste tube in the
other, align the tube nozzle above the bristles and squeeze out a bead
of toothpaste, and return the toothbrush upright into its holder and
the toothpaste back onto the table.

\textit{Grading rubric (additive):}
\begin{itemize}
    \item $+0.2$: \textbf{(a)} Grasps the toothbrush.
    \item $+0.1$: \textbf{(b)} Grasps the toothpaste tube.
    \item $+0.4$: \textbf{(c)} dispenses a bead of toothpaste onto the bristles.
    \item $+0.2$: \textbf{(d)} Returns the toothbrush upright into its holder.
    \item $+0.1$: \textbf{(e)} Places the toothpaste tube back on the table.
\end{itemize}


\begin{figure}[h]
    \centering
    \includegraphics[width=\linewidth]{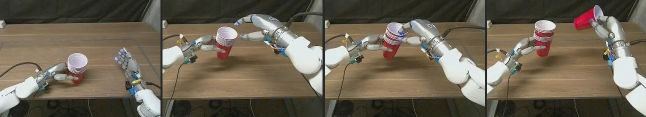}
    \caption{Key stages of Task~V: Split Cup.}
    \label{fig:split_cup}
\end{figure}

\noindent\textbf{Task V: Split Cup.} \textit{Text Instruction:}
``A stack of red plastic cups sits on the desktop; use the right hand to slide out the topmost one, exerting effort to separate it from the rest of the stack.'' Given a stack of nested cups on the
table, the robot must grasp the stack with the left hand to stabilize
it, and use the right hand to twist and rub exactly one cup off the
top of the stack.

\textit{Grading rubric (additive):}
\begin{itemize}
    \item $+0.2$: \textbf{(a)} Left hand grasps and stabilizes the cup stack.
    \item $+0.3$: \textbf{(b)} Right hand grasps the topmost cup of the stack.
    \item $+0.3$: \textbf{(c)} Right hand twists and separates exactly one cup
          from the stack.
    \item $+0.2$: \textbf{(d)} Right hand holds the single separated cup intact.
\end{itemize}

\begin{figure}[h]
    \centering
    \includegraphics[width=\linewidth]{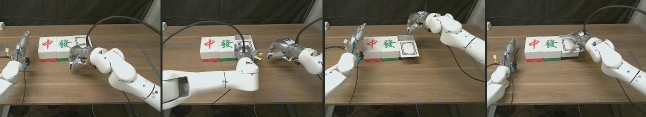}
    \caption{Key stages of Task~VI: Sort Mahjong.}
    \label{fig:sort_mahjong}
\end{figure}

\noindent\textbf{Task VI: Sort Mahjong.} \textit{Text Instruction:}
``Three boxes are placed on the table, representing the Mahjong tiles 'Red Zhong', 'Green Fa', and 'White Blank', respectively. In the center of the table lies a single Mahjong tile, placed face-down. Now, using your right hand, grasp the tile and discern its pattern; then, use your left hand to open the box corresponding to that pattern and place the tile inside.'' The robot must
pick up a face-down mahjong tile with the right hand and feel its
surface via tactile sensing to identify its category, then use the
left hand to slide open the lid of the matching compartment in the
organizer box, place the tile into the compartment with the right
hand, and close the lid with the right thumb.

\textit{Grading rubric (additive):}
\begin{itemize}
    \item $+0.1$: \textbf{(a)} Right hand picks up the face-back mahjong tile.
    \item $+0.5$: \textbf{(b)} Left hand slides open the lid of the correct
          compartment.
    \item $+0.2$: \textbf{(c)} Right hand places the tile into the correct
          compartment.
    \item $+0.2$: \textbf{(d)} Right thumb closes the compartment lid.
\end{itemize}

\begin{figure}[h]
    \centering
    \includegraphics[width=\linewidth]{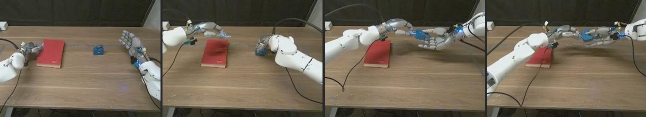}
    \caption{Key stages of Task~VII: Open Lock.}
    \label{fig:open_lock}
\end{figure}

\noindent\textbf{Task VII: Open Lock.} \textit{Text Instruction:}
``On the left side of the desk lies a red book, atop which rests a gray key; on the right side is a lock. Using your left thumb and index finger, slide the key free; then, pick up the lock with your right hand and use the key to unlock it.'' The robot must first grasp the
key with one hand and the padlock with the other, align and insert
the key into the keyhole, and rotate it to release the shackle.

\textit{Grading rubric (additive):}
\begin{itemize}
    \item $+0.2$: \textbf{(a)} Grasps the key.
    \item $+0.1$: \textbf{(b)} Grasps the padlock.
    \item $+0.4$: \textbf{(c)} Aligns and inserts the key into the keyhole.
    \item $+0.3$: \textbf{(d)} Rotates the key and successfully opens the lock.
\end{itemize}

\begin{figure}[h]
    \centering
    \includegraphics[width=\linewidth]{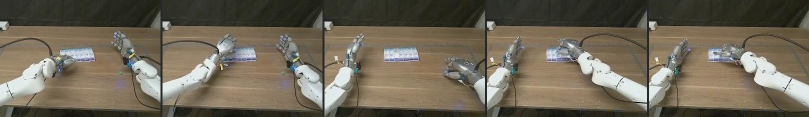}
    \caption{Key stages of Task~VIII: Refill Tablet.}
    \label{fig:refill_tablet}
\end{figure}

\noindent\textbf{Task VIII: Refill Tablet.} \textit{Text Instruction:}
``Use your left hand to open one of the compartments in the small box, use your right hand to grasp the small ball on the table, place the ball into the box, and then close the box.'' The robot must
use the left index finger to press the button on a compartment lid to
unlock it, flip the lid open with the left thumb, pick up the ball
with the right hand, place the ball into the open compartment, and
press the lid closed with the right index finger.

\textit{Grading rubric (additive):}
\begin{itemize}
    \item $+0.2$: \textbf{(a)} Left index finger presses the compartment button
          to unlock the lid.
    \item $+0.2$: \textbf{(b)} Left thumb flips the lid open.
    \item $+0.2$: \textbf{(c)} Right hand picks up the ball.
    \item $+0.2$: \textbf{(d)} Right hand places the ball into the open
          compartment.
    \item $+0.2$: \textbf{(e)} Right thumb presses the lid closed.
\end{itemize}


\begin{figure}[h]
    \centering
    \includegraphics[width=\linewidth]{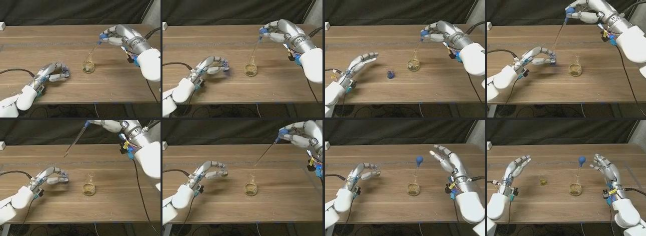}
    \caption{Key stages of Task~IX: Acid-Base Neutralization.}
    \label{fig:acid_base_neutralization}
\end{figure}

\noindent\textbf{Task IX: Acid-Base Neutralization.} \textit{Text Instruction:}
``On the right side of the desktop stands an Erlenmeyer flask containing 200 mL of citric acid solution; on the left is a beaker holding 20 mL of NaOH solution, which includes bromothymol blue indicator—appearing blue due to its alkaline nature. Using your right hand, pick up the dropper and draw up approximately 5 mL of the acid solution; then, using your left hand to hold the beaker, perform an acid-base titration until the liquid in the beaker turns green or yellow.'' The robot uses a dropper held
in the right hand to aspirate liquid from a conical flask, dispenses it
into a beaker held in the left hand, and swirls the beaker until the
blue indicator solution fully turns colorless, and then returns the dropper to the conical flask and places the beaker back on the table.

\textit{Grading rubric (additive):}
\begin{itemize}
    \item $+0.1$:  \textbf{(a)} Right hand grasps the dropper from the conical flask.
    \item $+0.15$: \textbf{(b)} Right hand aspirates liquid from the conical flask.
    \item $+0.1$:  \textbf{(c)} Left hand picks up the beaker.
    \item $+0.15$: \textbf{(d)} Right hand dispenses liquid from the dropper into the beaker.
    \item $+0.15$: \textbf{(e)} Left hand swirls the beaker to mix the contents.
    \item $+0.15$: \textbf{(f)} The solution in the beaker fully transitions from blue to colorless.
    \item $+0.1$:  \textbf{(g)} Right hand returns the dropper to the conical flask.
    \item $+0.1$:  \textbf{(h)} Left hand places the beaker back on the table.
\end{itemize}

\begin{figure}[h]
    \centering
    \includegraphics[width=\linewidth]{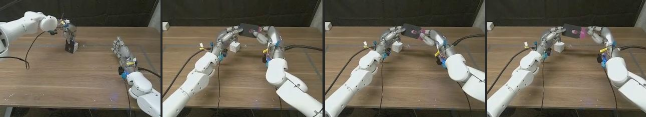}
    \caption{Key stages of Task~X: Extract Card.}
    \label{fig:extract_card}
\end{figure}

\noindent\textbf{Task X: Extract Card.} \textit{Text Instruction:}
``Next to the cube on the table lies a card case containing two cards. Pick up the case with the left hand, then use the right thumb to slide the cards out through the central opening; subsequently, use the right thumb and index finger to slide out the first card, taking care not to pull out the second one.'' The robot must pick up the
card sleeve (containing two cards) with the left hand, use the right
thumb to rub the cards partially out, then use the right thumb and
index finger to push the bottom card back in so that only the top
card remains exposed, and extract that single card.

\textit{Grading rubric (additive):}
\begin{itemize}
    \item $+0.2$: \textbf{(a)} Left hand picks up and holds the card sleeve.
    \item $+0.3$: \textbf{(b)} Right thumb rubs the cards partially out of the sleeve.
    \item $+0.3$: \textbf{(c)} Right thumb and index finger push the bottom card in.
    \item $+0.2$: \textbf{(d)} Right hand extracts the single top card from the sleeve.
\end{itemize}

\begin{figure}[h]
    \centering
    \includegraphics[width=\linewidth]{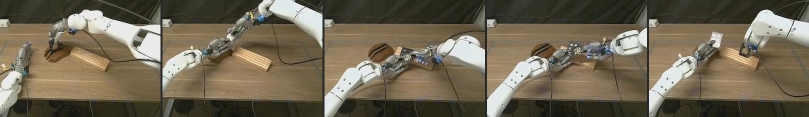}
    \caption{Key stages of Task~XI: Deal Poker.}
    \label{fig:deal_poker}
\end{figure}

\noindent\textbf{Task XI: Deal Poker.} \textit{Text Instruction:}
``Pick up a stack of playing cards with your right hand, then transfer it to your left; hold the stack aloft with your left hand, use your right thumb to slide out the top card, grasp it, and place it into the card holder.'' The robot must
grasp the full card stack from above with the right hand, transfer it
to the left hand, use the right thumb to flick the top card partially
out, adjust with the right thumb and index finger until exactly one
card protrudes, grasp that single card, and insert it vertically into
the dedicated card slot.

\textit{Grading rubric (additive):}
\begin{itemize}
    \item $+0.1$: \textbf{(a)} Right hand grasps the card stack from above.
    \item $+0.2$: \textbf{(b)} Right hand transfers the stack to the left hand
          (handover).
    \item $+0.3$: \textbf{(c)} Right thumb flicks and adjusts until exactly one
          card protrudes from the top.
    \item $+0.3$: \textbf{(d)} Right hand successfully grasps the single protruding card.
    \item $+0.1$: \textbf{(e)} Right hand inserts the card vertically into the
          card slot.
\end{itemize}

\begin{figure}[h]
    \centering
    \includegraphics[width=\linewidth]{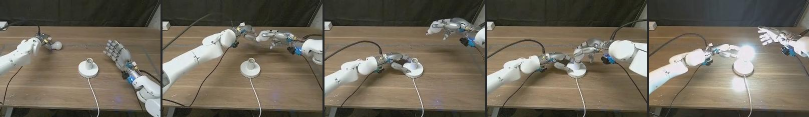}
    \caption{Key stages of Task~XII: Screw Lightbulb.}
    \label{fig:screw_lightbulb}
\end{figure}

\noindent\textbf{Task XII: Screw Lightbulb.} \textit{Text Instruction:}
``There is a lightbulb and a base on the desktop. Use your left hand to pick up the lightbulb and transfer it to your right hand; then, use your left hand to hold down the base while using your right hand to screw the lightbulb into the base until it lights up.'' The robot must pick up the
lightbulb with the left hand, transfer it to the right hand (handover),
stabilize the lamp socket with the left hand, and use the right hand
to rotate the bulb through multiple turns into the socket until it
is fully seated and illuminates.

\textit{Grading rubric (additive):}
\begin{itemize}
    \item $+0.1$:  \textbf{(a)} Left hand picks up the lightbulb.
    \item $+0.2$: \textbf{(b)} Left hand transfers the bulb to the right hand
           (handover).
    \item $+0.1$:  \textbf{(c)} Left hand stabilizes the lamp socket.
    \item $+0.4$:  \textbf{(d)} Right hand aligns the bulb and rotates continuously to engage the threads.
    \item $+0.2$: \textbf{(e)} Bulb is fully seated and the lamp illuminates.
\end{itemize}

%
%

\section{T-Rex Dataset}
\label{apdx:trex_dataset}

The T-Rex Dataset is constructed to support large-scale mid-training of tactile-reactive dexterous manipulation policies. In the following, we describe the modalities recorded per episode, the object--motor-primitive taxonomy, the scene-level diversity, the quality-control pipeline, the language-annotation procedure, and the dataset's licensing and ethical considerations.

\minisection{Recorded Modalities and Episode Schema}
Each demonstration episode is stored as a time-aligned bundle of synchronized streams collected through the teleoperation stack described in Fig.~\ref{fig:robotsys}. Specifically, every episode contains: (i) three monocular RGB streams (one head ZED~X~Mini and two wrist-mounted ZED~X~One~S wide-view cameras) at $640\times360$ resolution and 30~Hz; (ii) bimanual proprioception consisting of $2\times7$ arm joint positions and velocities together with the $2\times22$-DoF Sharpa Wave hand joint states; (iii) $SE(3)$ end-effector poses of both wrists; (iv) per-fingertip tactile observations for all ten fingertips, comprising a single-channel deformation depth map and a 6-axis net wrench; and (v) the natural-language task instruction associated with the episode (see \textit{``Automated VLM-based Language Annotation''} below). All streams share a common timestamp and are recorded at the 30~Hz cadence of the high-level teleoperation thread, ensuring tight temporal alignment between vision, proprioception, action, and tactile signals.

\minisection{Data Taxonomy}
To ensure broad coverage of contact-rich manipulation behaviors, we construct the dataset taxonomy by systematically combining 207 common household objects with 22 motor primitives and retaining only physically feasible object--motor primitive pairs. Out of the $207\times22=4{,}554$ candidate combinations, infeasible pairs (e.g., the \emph{pour} primitive applied to a solid block, or the \emph{twist} primitive applied to a non-articulated object) are pruned via a per-primitive feasibility checklist annotated manually. This process yields 502 unique object--motor primitive combinations, comprising 7755 episodes and 100 hours of demonstrations, with a median episode length of 29.8~s and an interquartile range of 21.0--41.1~s. Each retained pair receives on average $\sim$16 demonstrations to expose the policy to the full action distribution of every primitive applied to every compatible object. Demonstrations were collected by  teleoperators over a period of 10 weekss. The resulting distribution of object categories, motor primitives, and object--motor primitive pairs is shown in Fig.~\ref{fig:play_data_diversity}.

\minisection{Scene Diversity}
To improve visual robustness and support language-conditioned behavior, we collect data under diverse scene configurations. Specifically, we use six distinct tabletop backdrops and vary the arrangement of surrounding objects across demonstrations. During data collection, randomly selected distractor objects (drawn from a pool of more than 210 non-target items, with typically 0--5 distractors visible per scene) are placed alongside the target object to increase scene complexity and encourage the policy to identify and manipulate the correct object based on task context and language instructions. Furthermore, for each object--motor skill pair, we randomize the initial object position and orientation at the start of every episode. Combined with the large variety of objects and motor primitives, these variations expose the policy to substantial visual and spatial diversity, reducing overfitting to specific scene layouts and improving generalization to unseen environments.

\minisection{Data Cleaning}
After data collection, we perform a data-cleaning stage to ensure the quality and consistency of the dataset. We remove episodes containing unstable tactile measurements, corrupted sensor streams, or abnormal motions caused by teleop failures. We further filter demonstrations exhibiting extreme joint-space velocities or other artifacts that may negatively affect policy learning.

\minisection{Automated VLM-based Language Annotation Baseline}
To scalably generate language instruction annotations across diverse tasks, we annotate each episode with a commercial vision--language foundation model. For every episode, we feed the model a set of sampled image frames (subsampled 4 to 6 frames from the head camera view) together with the minimal labels recorded during teleoperation (target object name and motor-primitive name), and prompt the model to compose a single imperative sentence that comprihensively describes the episode's motion. The resulting annotations are then verified by human annotators to filter out hallucinations and imprecise descriptions.

\minisection{Ethical Considerations and Dataset Release}
All T-Rex demonstrations are collected in a controlled laboratory environment using the Dexmate Vega-1 research platform; no third-party human subjects appear in the released RGB streams, and incidental frames containing teleoperator hands during reset interactions are clipped from the released episodes. The household objects used during data collection are commodity items that contain no personally identifying information. Teleoperators provided informed consent for the recording and release of their teleoperation data. We plan to release the T-Rex dataset, including raw sensor streams, derived tactile representations, and language annotations, under the MIT license, together with the data loaders and pre-processing scripts required to reproduce the results in this paper.

\section{Failure Case Analysis}
\label{apdx:fail_case}

\begin{figure}[!t]
    \centering
    \includegraphics[width=\linewidth]{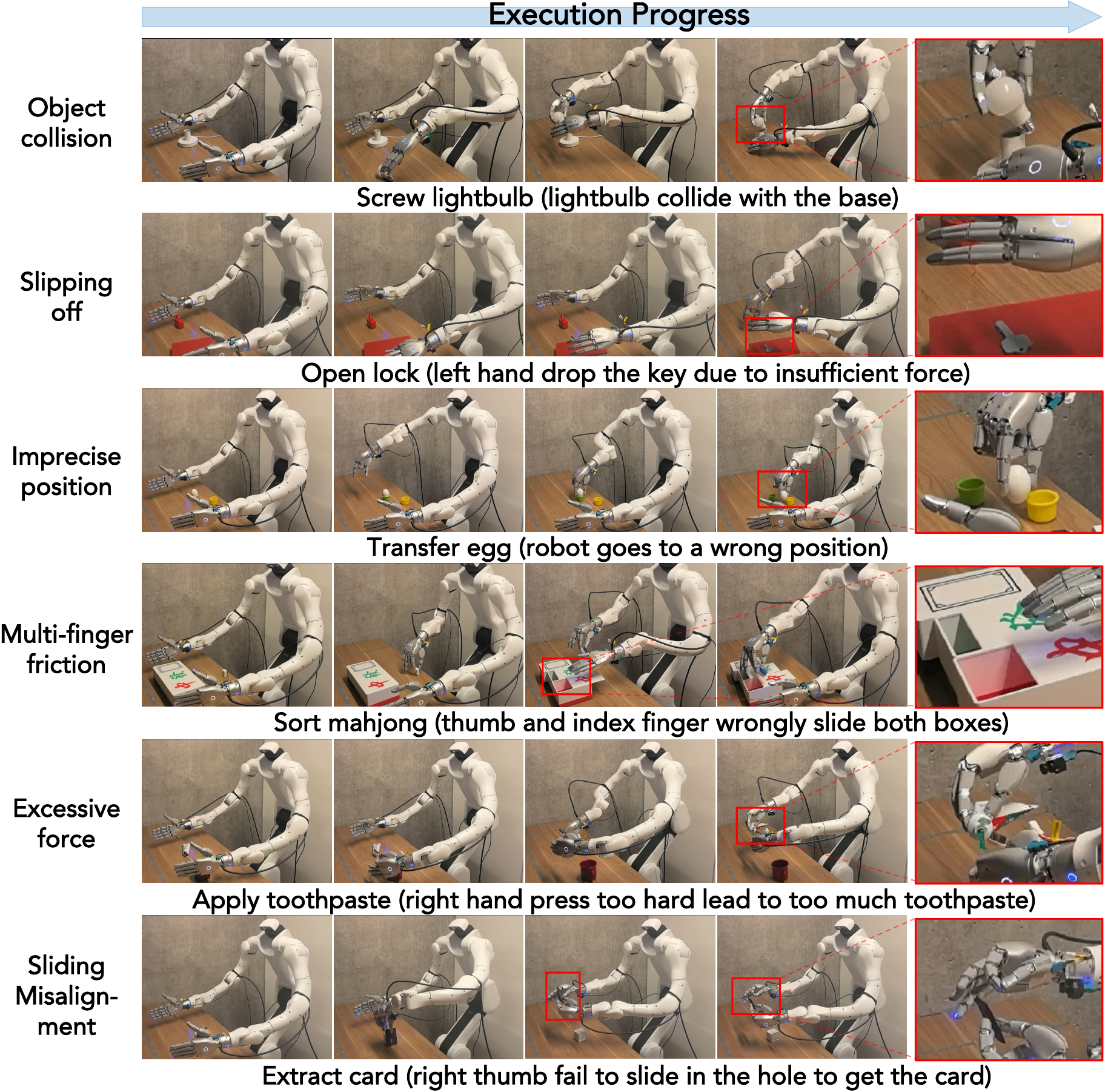}
    \caption{\textbf{Failure Case Analysis.} The order from left to right indicates the execution progress of the tasks, while the final column illustrates the specific failure scenarios.}
    \label{fig:fail_case}
\end{figure}

Across various scenarios and tasks, we observed a diverse range of failure cases, as illustrated in Fig.~\ref{fig:fail_case}; specifically, the red boxes highlight the contact issues that occurred during these failures.

\textbf{1) Object Collision.} During the \textit{screw lightbulb} task in the first row, the right hand failed to correctly insert it into the socket after grasping the lightbulb; instead, it caused the lightbulb to collide with the base, thereby preventing the subsequent insertion and rotation steps from being completed. This indicates that during the execution of complex tasks, there remains a lack of fine-grained visual alignment, and that excessively rapid motion execution can lead to object collisions.

\textbf{2) Slipping Off.} During the \textit{open lock} task in the second row, the model successfully slid and grasped the key; However, it failed to maintain a secure grip during the subsequent steps, causing the key to slip and drop. For the grasping of small objects and precise in-hand manipulation, the model still lacks a certain degree of fine-grained dexterity, which remains a limitation attributable to the data distribution of the teleoperated data.

\textbf{3) Imprecise Position.} In the task of \textit{transfer egg}, the model successfully grasped the eggs and relied on force feedback to ensure its integrity. But it failed to place the egg correctly into the yellow egg tray. This demonstrates that the model still suffers from deficiencies in precise positioning, which is a limitation that highlights the inherent distribution shift characteristic of Behavioral Cloning (BC).

\textbf{4) Multi-finger friction.} In the \textit{sort mahjong} task, the model correctly selected the "Red Zhong" tile located on the left as the target box to be opened; however, the positioning of its thumb was too low, causing it to make contact with the central "Green Fa" tile and inadvertently open two boxes simultaneously. This highlights that dexterous hand control still lacks coordination at the individual finger level, and issues such as unintended contact between multiple fingers may persist.

\textbf{5) Excessive Force.} During the \textit{apply toothpaste} task, after grasping the tube, the model applied excessive force and squeezed out too much toothpaste, resulting in a failure to catch it with the toothbrush. This highlights that in the manipulation of certain deformable objects, the model remains constrained by the overly forceful control inherent in its sequencial prediction mechanism.

\textbf{6) Sliding Misalignment.} In the \textit{extract card} task, after grasping the card sleeve, the model failed to apply uniform force when extracting the card from the small slot; this suggests that for tasks requiring sliding motions, the model needs to establish stronger tactile conditioning in the temporal dimension to generate the correct actions.

\end{document}